\documentclass[sigconf]{acmart}
\AtBeginDocument{%
  }

\setcopyright{acmlicensed}
\copyrightyear{2018}
\acmYear{2018}
\acmDOI{XXXXXXX.XXXXXXX}
\acmConference[Conference acronym 'XX]{Make sure to enter the correct
  conference title from your rights confirmation emai}{June 03--05,
  2018}{Woodstock, NY}
\acmISBN{978-1-4503-XXXX-X/18/06}

\usepackage{mdframed}
\usepackage{enumitem}
\usepackage{tikz}
\usetikzlibrary{arrows.meta, positioning}
\usetikzlibrary{graphs,quotes} 
\usepackage{listings}

\begin{document}

\title{OLG++: A Semantic Extension of Obligation Logic Graphs}

\author{Subhasis Dasgupta}
\affiliation{%
  \institution{University of California San Diego}
  \city{La Jolla}
  \state{California}
  \country{USA}
}
\email{sudasgupta@ucsd.edu}

\author{Jon Stephens}
\affiliation{%
  \institution{University of California San Diego}
  \city{La Jolla}
  \state{California}
  \country{USA}
}
\email{jstephe@ucsd.edu}

\author{Amarnath Gupta}
\affiliation{%
  \institution{University of California San Diego}
  \city{La Jolla}
  \state{California}
  \country{USA}
}
\email{a1gupta@ucsd.edu}

\renewcommand{\shortauthors}{Dasgupta et al.}

\begin{abstract}
We present OLG++, a semantic extension of the Obligation Logic Graph (OLG) for modeling regulatory and legal rules in municipal and interjurisdictional contexts. OLG++ introduces richer node and edge types, including spatial, temporal, party group, defeasibility, and logical grouping constructs, enabling nuanced representations of legal obligations, exceptions, and hierarchies. The model supports structured representation of rules with contextual conditions, precedence, and complex triggers. We demonstrate its use through examples from food-business regulations, showing how OLG++ supports legal question answering using property-graph queries. We also discuss how OLG++ can complement LegalRuleML by providing graph-native constructs for subclass relations, spatial constraints, and reified exception structures. The worked examples and first-pass coverage analysis show that, on the dimensions studied, OLG++ is more expressive than the baseline OLG model for municipal regulatory representation.
\end{abstract}


\maketitle

\section{Introduction}

Regulatory compliance increasingly requires systems that can answer questions over legal texts whose meaning depends on obligations, permissions, prohibitions, exceptions, temporal constraints, locations, jurisdictional scope, and regulated activities. This problem is especially visible in municipal and regional regulation. A food truck business, for example, may prepare food in one jurisdiction, store equipment in another, and sell products in a third. Determining which rules apply may require combining health-code obligations, vending restrictions, permit requirements, school-zone or park-zone restrictions, time-of-day rules, and exceptions for emergencies or special events. Such questions are difficult to answer from text alone because the relevant facts are distributed across clauses, definitions, cross-references, and jurisdictional boundaries.

Knowledge representation provides one way to make such legal content explicit and queryable. In the legal domain, prior work has developed XML-based, rule-based, and graph-based representations for legal documents and legal reasoning, including LegalRuleML and related approaches to legal knowledge interchange and legal question answering \cite{boer2008metalex,athan2015legalruleml,legalruleml-core-1.0,filtz2021knowledge,martinez2023survey}. Large-scale legal semantic platforms such as Lynx further illustrate the role of legal knowledge graphs and semantic enrichment services for legal documents~\cite{schneider2022lynx}. These representations differ in their goals. Some emphasize interchange and normative rule expression; others emphasize search, retrieval, or reasoning over structured legal knowledge. Our goal in this paper is narrower and complementary: we study how a graph representation of obligations can be extended so that regulatory provisions involving locations, jurisdictions, defeasible rules, party structures, and nested conditions can be represented at a level suitable for graph querying and downstream question answering.

Obligation Logic Graphs (OLG) \cite{servantez2023computable} provide a useful starting point. OLG was originally developed for computational contracts and is designed to extract the deontic content of contractual clauses, including obligations, prohibitions, permissions, parties, actions, temporal expressions, amounts, references, and formulaic dependencies. This makes OLG attractive for legal knowledge representation beyond contracts: statutes, ordinances, and regulations also contain deontic content. However, municipal and regulatory texts introduce additional structure that is less central in many contract settings. Rules may apply only within particular jurisdictions, within specified distances of schools or parks, during recurring time intervals, or under nested combinations of conditions. They may also include exceptions and overrides that depend on permits, emergencies, or more specific local rules.

This paper introduces OLG++, a semantic extension of OLG designed for regulatory and municipal legal texts. OLG++ should be understood as a representation layer, not as a replacement for foundational ontologies, geospatial ontologies, or legal-rule interchange standards. Domain concepts such as \textit{food truck}, \textit{sidewalk}, \textit{school}, \textit{commercial kitchen}, and \textit{public park} may be imported from, or aligned with, external legal and domain ontologies, including legal core ontologies such as LKIF-Core~\cite{hoekstra2007lkif}. OLG++ provides the graph-level legal structure that connects such entities to deontic triggers, parties, actions, conditions, temporal expressions, locations, jurisdictions, references, exceptions, and precedence relationships. Thus, OLG++ is intended to complement external ontologies rather than duplicate them.

The paper also separates representation from extraction. We do not claim that legal text can be converted automatically into OLG++ without linguistic analysis or human validation. Extracting OLG++ graphs from statutes and ordinances is a separate NLP and information-extraction problem involving clause segmentation, definition recognition, semantic typing, relation extraction, source-span alignment, and normalization against external vocabularies. The present paper focuses on the representation question: assuming that relevant spans and entities can be identified, what graph schema is needed to faithfully represent the legal structure required for regulatory question answering? This separation allows us to analyze the expressivity of the representation itself while leaving a full extraction pipeline for future work.

Our contributions are as follows. First, we analyze the original OLG model using examples from food-business and municipal regulation, identifying representational gaps involving location, jurisdiction, logical grouping, semantic granularity, party groups, and defeasibility. Second, we define a normalized OLG++ schema that extends OLG with explicit node and edge types for locations, specific locations, jurisdictions, semantic entities, party groups, logical groups, condition groups, exceptions, overrides, precedence, and spatial and temporal relationships. Third, we clarify the intended semantics of constructs that are often ambiguous in informal graph models, including the distinction between an action node and a \texttt{whatRel} edge, the difference between exceptions and overrides, and the role of party groups. Fourth, we present case-study queries, written in Cypher \cite{francis2018cypher}, showing how OLG++ can support graph queries that combine deontic, spatial, temporal, jurisdictional, and defeasible information.

The remainder of the paper is organized as follows. Section~\ref{sec:OLG} reviews the original OLG model and illustrates its limitations on regulatory examples. Section~\ref{sec:OLG++} presents the normalized OLG++ schema and explains the semantics of the added constructs. Section~\ref{sec:using-olgpp} gives worked OLG++ representations and case-study query patterns over those representations. Section~\ref{sec:evaluation} presents a preliminary coverage analysis over selected Carlsbad Municipal Code provisions. Section~\ref{sec:discussion-conclusion} discusses the relationship between OLG++, external ontologies, LegalRuleML, extraction pipelines, and future work.

\begin{table*}
\caption{Core OLG schema used as the baseline in this paper. The original OLG terminology uses \textit{what} both for a node type and for a relationship. We use \texttt{whatRel} for the relationship to avoid this ambiguity.}
\label{tab:olg_core_schema}
\centering
\begin{tabular}{|l|p{4.3cm}|p{8.2cm}|}
\hline
\multicolumn{1}{|c|}{\textbf{Category}} &
\multicolumn{1}{c|}{\textbf{Node or Edge Type}} &
\multicolumn{1}{c|}{\textbf{Intended Meaning}} \\
\hline
\multicolumn{3}{|c|}{\textbf{Node Types}} \\
\hline
Trigger &
\texttt{obligation\_trigger} &
A textual trigger that introduces an obligation, prohibition, permission, or entitlement. Examples include `must maintain,'' `may sell,'' and `shall not park.'' \\
\cline{2-3}
&
\texttt{event\_trigger} &
An event or condition whose occurrence may trigger, defeat, or modify an obligation. Examples include `request by a purchaser,'' `inspection failure,'' or `late payment.'' \\
\hline
Party &
\texttt{party} &
A legal actor or participant to whom a deontic modality may attach, such as a vendor, purchaser, agency, owner, manager, or inspector. \\
\hline
Action/Object &
\texttt{what} &
The regulated action, object, or state of affairs that is the subject of an obligation, prohibition, permission, or entitlement. Examples include selling food, parking a truck, maintaining a log, or storing equipment. \\
\hline
Temporal &
\texttt{timex} &
A temporal expression, duration, deadline, interval, or recurrence pattern. Examples include `60 minutes,'' `during business hours,'' and `every day.'' \\
\hline
Reference &
\texttt{reference} &
A legal or documentary source on which a rule depends, such as a statute, ordinance, section, clause, permit, or incorporated standard. \\
\hline
Amount &
\texttt{amount} &
A numerical quantity, threshold, fee, distance, count, or measurement. Examples include `500 feet,'' `\$500,'' `five tables,'' and `41°F.'' \\
\hline
\multicolumn{3}{|c|}{\textbf{Edge Types}} \\
\hline
Deontic Modality &
\texttt{obligation}, \texttt{prohibition}, \texttt{permission}, \texttt{entitlement} &
Links an \texttt{obligation\_trigger} to the party to whom the modality applies. For example, a permission edge may link the trigger `may sell'' to the party `food truck vendor.'' \\
\hline
Regulated Subject &
\texttt{whatRel} &
Links a trigger to the regulated action, object, or state of affairs. For example, `may sell'' is linked by \texttt{whatRel} to the action ``sell food.'' \\
\hline
Logical Dependency &
\texttt{if\_true}, \texttt{if\_false}, \texttt{if\_late} &
Links an event or condition to a consequent obligation depending on whether the antecedent is true, false, or late. \\
\hline
Subject To &
\texttt{subject\_to} &
Links a trigger or regulated action to a legal reference or external clause. \\
\hline
Temporal &
\texttt{before}, \texttt{after}, \texttt{on}, \texttt{during}, \texttt{recurring} &
Links a trigger, event, or action to a temporal expression. \\
\hline
Amount &
\texttt{amount} &
Links a trigger, action, or formula to a numerical quantity. \\
\hline
Formula &
\texttt{addition}, \texttt{subtraction}, \texttt{multiplication}, \texttt{division}, \texttt{maximum}, \texttt{minimum} &
Represents mathematical or logical operations between quantities or formula components. \\
\hline
Logical Operator &
\texttt{and}, \texttt{or} &
Connects logical dependencies or conditions to form compound conditions. In the baseline OLG model, these operators are available as relationships, but complex nested logical groupings are not first-class nodes. \\
\hline
\end{tabular}
\end{table*}

\section{OLG as a Baseline}
\label{sec:OLG}

The Obligation Logic Graph (OLG) model provides a compact graph representation for the deontic content of legal and contractual language. Its central idea is that legal text can be decomposed into triggers, parties, regulated actions or objects, temporal expressions, references, amounts, and formulaic dependencies. Table~\ref{tab:olg_core_schema} gives the baseline schema used in this paper. We use the term \texttt{whatRel} for the edge connecting a deontic trigger to its regulated subject because the original OLG terminology uses \textit{what} both as a node type and as a relationship. Thus, in this paper, \texttt{what} is a node type, while \texttt{whatRel} is an edge type.

\subsection{A Minimal Regulatory Example}

We first use a deliberately simple municipal rule to illustrate the baseline representation:

\begin{quote}
Food truck vendors may sell food from a food truck only at the request of a bona fide purchaser.
\end{quote}

This rule contains a permission, a party who receives that permission, a regulated action, and a condition under which the permission applies. The trigger is the phrase `may sell,'' the party is `food truck vendor,'' the regulated action is `sell food from a food truck,'' and the event condition is `request of a bona fide purchaser.'' Table~\ref{tab:olg_simple_example_nodes} shows the corresponding OLG nodes.

\begin{table}[t]
\caption{OLG nodes for the minimal food-truck permission example.}
\label{tab:olg_simple_example_nodes}
\centering
\resizebox{\columnwidth}{!}{%
\begin{tabular}{|l|l|p{6.2cm}|}
\hline
\textbf{Node ID} & \textbf{Type} & \textbf{Label / Description} \\
\hline
$n_1$ & \texttt{obligation\_trigger} &
Permission trigger: ``may sell.'' \\
\hline
$n_2$ & \texttt{party} &
Food truck vendor. \\
\hline
$n_3$ & \texttt{what} &
Sell food from a food truck. \\
\hline
$n_4$ & \texttt{event\_trigger} &
Request by a bona fide purchaser. \\
\hline
\end{tabular}
}
\end{table}

The corresponding edges are straightforward: $(n_1,n_2)$ has modality \texttt{permission}; $(n_1,n_3)$ is a \texttt{whatRel} edge; and $(n_4,n_1)$ is an \texttt{if\_true} edge stating that the permission applies only if the purchaser-request event occurs. This example shows the strength of OLG: simple deontic rules can be represented in a small graph whose nodes and edges have a clear legal interpretation.

\begin{figure}[t]
\centering
\resizebox{\columnwidth}{!}{%
\begin{tikzpicture}[
  node/.style={circle, draw, text centered, minimum size=0.7cm, font=\sffamily\small},
  edgeLabel/.style={draw=none, fill=white, inner sep=1pt, font=\scriptsize},
  >=Latex
]
\node[node, fill=blue!20] (n1) at (0,0) {$n_1$};
\node[node, fill=green!20] (n2) at (-2.2,-1.6) {$n_2$};
\node[node, fill=red!20] (n3) at (0,-1.6) {$n_3$};
\node[node, fill=yellow!30] (n4) at (2.2,-1.6) {$n_4$};

\draw[->] (n1) -- node[edgeLabel, above left] {\texttt{permission}} (n2);
\draw[->] (n1) -- node[edgeLabel, right] {\texttt{whatRel}} (n3);
\draw[->] (n4) -- node[edgeLabel, above right] {\texttt{if\_true}} (n1);
\end{tikzpicture}%
}
\caption{A minimal OLG representation. The permission trigger $n_1$ is linked to the party $n_2$, the regulated action $n_3$, and the purchaser-request event $n_4$.}
\label{fig:olg_minimal_example}
\end{figure}

\subsection{Where the Baseline Model Begins to Strain}

The limitations of the baseline model become clearer when we move from this simple permission to a more typical municipal rule:

\begin{quote}
A food truck vendor may not park or stand for longer than 60 minutes on any public street, alley, highway, parking lot, sidewalk, or right-of-way.
\end{quote}

The deontic structure is still recognizable. The trigger is a prohibition, the party is a food truck vendor, the regulated action is parking or standing, and the temporal condition is a duration of 60 minutes. However, the locational expression is more complex. It enumerates several distinct location classes: public street, alley, highway, parking lot, sidewalk, and right-of-way. If these are collapsed into a single \texttt{what} node labeled `public locations,'' then the graph loses semantic granularity. A later query such as `what restrictions apply to sidewalks?'' would require string search or additional text processing inside the node label rather than graph traversal over explicit location nodes.

This example also shows that the temporal and locational restrictions have different logical roles. The 60-minute duration is a temporal condition. The list of public locations is a disjunction: the rule applies on a street, or an alley, or a highway, or a parking lot, or a sidewalk, or a right-of-way. The overall applicability condition combines the temporal condition and the locational condition conjunctively. Informally, the prohibition applies when the vendor parks or stands for more than 60 minutes \textit{and} the location is one of the enumerated public-location types. The baseline OLG model has \texttt{and} and \texttt{or} relationships, but it does not provide a first-class representation of nested logical groupings such as:\\
$\textit{duration} > 60\textit{ minutes~}
\wedge
(\textit{street} \vee \textit{alley} \vee \textit{highway~} \vee
\textit{parking lot} \vee \textit{sidewalk} \vee \textit{right-of-way})$\\
Without a logical-group node, the graph must either flatten the condition or encode part of the structure informally in labels and attributes.

\subsection{Representational Gaps}

These examples suggest four limitations of the baseline OLG model for municipal and regulatory texts.

First, OLG does not distinguish sufficiently between regulated actions and domain entities. The \texttt{what} node can represent an action such as selling, parking, or storing, but it may also be used to hold a complex noun phrase or an enumerated list. This makes the graph easy to construct but weakens queryability. In regulatory settings, terms such as \textit{sidewalk}, \textit{school}, \textit{commercial kitchen}, \textit{freezer}, \textit{thermometer}, and \textit{raw chicken} should be represented as typed domain entities rather than embedded in long node labels.

Second, OLG has limited support for locational and jurisdictional structure. Municipal regulations frequently refer to city limits, zoning districts, residential zones, public rights-of-way, schools, parks, beaches, and named facilities. These locations may stand in topological, distance-based, or jurisdictional relationships: one area may be within another, a vending location may be within 500 feet of a school, and a local rule may apply only within a particular city or district. Treating these expressions as ordinary \texttt{what} nodes obscures their spatial semantics.

Third, OLG represents logical dependencies but does not provide an explicit structure for nested logical groupings. Regulatory provisions often combine conditions using mixtures of conjunction, disjunction, and negation. A food-truck rule may apply if the operation occurs during daylight hours and is at least 500 feet from a school, or if the operation is part of a permitted neighborhood event and has written approval from affected residents. Such rules require explicit condition groups and expression trees, not only binary \texttt{and} and \texttt{or} edges.

Fourth, OLG does not directly model rule defeasibility at the level needed for statutory and municipal law. Regulations often include exceptions, overrides, and priority relations. A general prohibition may be defeated by a special-event permit; an emergency rule may override an ordinary permission; and a more specific jurisdictional rule may narrow a more general citywide rule. These relations are not merely additional conditions on a single rule. They define how rules interact when multiple provisions are applicable.

The goal of OLG++ is to preserve the compact deontic structure of OLG while adding the semantic machinery needed for regulatory question answering: typed domain entities, explicit location and jurisdiction nodes, logical and condition groups, party groups, and defeasibility relations. The next section introduces the normalized OLG++ schema.

\section{OLG++: A Normalized Semantic Extension of OLG}
\label{sec:OLG++}

The preceding section showed that the baseline OLG model is useful for representing simple deontic structures, but becomes under-specified when applied to municipal and regulatory provisions involving locations, jurisdictions, nested conditions, regulated objects, collective parties, and defeasible rule interactions. OLG++ extends OLG as a typed property-graph schema for such settings. It preserves the core OLG idea that legal provisions can be decomposed into deontic triggers, parties, actions, temporal expressions, references, amounts, and formulaic relationships, but it adds semantic typing and integrity constraints so that regulatory rules can be queried at the level of explicit graph structure rather than long textual labels.

OLG++ is not intended to be a foundational ontology, a geospatial ontology, or a complete legal reasoning language. Rather, it is a representation layer for regulatory provisions. It connects deontic content to domain entities, locations, jurisdictions, condition structures, legal sources, and defeasibility relations. Domain concepts such as \textit{food truck}, \textit{sidewalk}, \textit{school}, \textit{commercial kitchen}, \textit{raw chicken}, and \textit{thermometer} may be imported from or aligned with external vocabularies. OLG++ therefore provides graph-level legal structure while allowing taxonomic and domain-specific knowledge to be supplied by external ontologies where available.

\subsection{Property Graph Assumptions and Naming Conventions}

OLG++ is based on the property graph model \cite{angles2018property}. Nodes and edges have labels, and both may carry key-value properties. We assume that every node has the base properties
\[
\texttt{id}, \texttt{created\_date}, \texttt{modified\_date}, \texttt{status},
\]
and that every edge has the base properties
\[
\texttt{id}, \texttt{creation\_date}, \texttt{modified\_date}, \texttt{temporal\_validity}.
\]
The \texttt{status} property is used because legal rules and derived graph fragments may be active, superseded, deprecated, or under review. The \texttt{temporal\_validity} property on edges is used because relationships between legal entities may change over time, even when the participating nodes remain stable.

For clarity, we use PascalCase for node labels and upper snake case for edge labels. Thus, the baseline OLG node type \texttt{obligation\_trigger} becomes \texttt{ObligationTrigger}, while the relation previously called \texttt{whatRel} becomes \texttt{WHAT\_REL}. We keep the term \texttt{WHAT\_REL} because the original OLG terminology uses \textit{what} both as a node type and as a relation. In OLG++, \texttt{What} is the node representing the regulated action, object, or state of affairs, while \texttt{WHAT\_REL} is the edge linking a trigger to that regulated subject.

\subsection{Normalized Node Schema}

Table~\ref{tab:olgpp_node_schema} gives the normalized OLG++ node schema. The schema includes both inherited OLG node types and OLG++ additions. The inherited node types are retained because they capture the basic deontic content of a provision. The added node types address semantic gaps encountered in regulatory settings.

\begin{table*}
\caption{Normalized OLG++ node schema. All nodes also have base properties \texttt{id}, \texttt{created\_date}, \texttt{modified\_date}, and \texttt{status}.}
\label{tab:olgpp_node_schema}
\centering
\small
\begin{tabular}{|p{3.0cm}|p{4.7cm}|p{4.7cm}|p{4.2cm}|}
\hline
\textbf{Node Label} & \textbf{Meaning} & \textbf{Important Properties} & \textbf{Examples} \\
\hline
\texttt{ObligationTrigger} &
A trigger introducing an obligation, prohibition, permission, or entitlement. &
\texttt{modality}, \texttt{description}, \texttt{priority}, \texttt{temporal\_scope}, \texttt{jurisdiction}, \texttt{source\_text}. &
`may sell,'' `shall not park,'' ``must maintain log.'' \\
\hline
\texttt{EventTrigger} &
An event whose occurrence activates, modifies, or defeats a rule. &
\texttt{event\_type}, \texttt{conditions}, \texttt{description}, \texttt{trigger\_status}, \texttt{source\_text}. &
Inspection failure; emergency response activated; purchaser request. \\
\hline
\texttt{Party} &
An atomic legal actor or participant. &
\texttt{name}, \texttt{role}, \texttt{party\_type}, \texttt{legal\_capacity}, \texttt{jurisdiction}. &
Food truck vendor; purchaser; health inspector; city manager. \\
\hline
\texttt{PartyGroup} &
A group of parties treated jointly, collectively, or by legal reference. &
\texttt{group\_type}, \texttt{is\_collective}, \texttt{liability\_mode}, \texttt{legal\_person\_status}. &
Food truck operator and kitchen manager; employees of a restaurant. \\
\hline
\texttt{What} &
The regulated action, object, or state of affairs governed by a rule. &
\texttt{description}, \texttt{category}, \texttt{scope}, \texttt{action\_type}, \texttt{object\_type}. &
Sell food; park vehicle; maintain inspection log; store food. \\
\hline
\texttt{SemanticEntity} &
Abstract superclass for non-party domain entities that occur in rules. &
\texttt{name}, \texttt{semantic\_type}, \texttt{category}, \texttt{external\_ontology\_id}. &
Equipment; food item; permit; vehicle; facility. \\
\hline
\texttt{Timex} &
Temporal expression, duration, interval, deadline, or recurrence. &
\texttt{temporal\_type}, \texttt{start}, \texttt{end}, \texttt{duration}, \texttt{recurrence}, \texttt{timezone}. &
60 minutes; every day; 2:00 a.m.; during business hours. \\
\hline
\texttt{Amount} &
Numeric quantity, threshold, fee, count, distance, or measurement. &
\texttt{value}, \texttt{unit}, \texttt{currency}, \texttt{quantity\_type}, \texttt{comparator}. &
500 feet; \$500; five tables; $41^{\circ}\mathrm{F}$. \\
\hline
\texttt{Reference} &
Legal source, clause, section, incorporated standard, or permit reference. &
\texttt{source}, \texttt{clause}, \texttt{section}, \texttt{jurisdiction}, \texttt{citation}, \texttt{uri}. &
Health Code \S114.2(b); vending permit; municipal code section. \\
\hline
\texttt{Location} &
General location class or spatial entity. &
\texttt{location\_type}, \texttt{description}, \texttt{boundary}, \texttt{geojson}. &
Sidewalk; public park; residential zone; commercial kitchen. \\
\hline
\texttt{SpecificLocation} &
Named place, address, or concrete spatial region. &
\texttt{name}, \texttt{address}, \texttt{lat}, \texttt{long}, \texttt{geojson}. &
2175 El Camino Real; San Diego Airport; Balboa Park. \\
\hline
\texttt{Jurisdiction} &
Spatial or institutional scope in which a rule applies. &
\texttt{name}, \texttt{jurisdiction\_type}, \texttt{boundary}, \texttt{parent\_jurisdiction}. &
San Diego; Tijuana; Central Business District. \\
\hline
\texttt{LocationPredicate} &
Reified spatial predicate or spatial expression. &
\texttt{predicate\_type}, \texttt{parameters}, \texttt{expression}, \texttt{normalized\_form} &
within(San Diego); northOf(Chula Vista); outside(school zone). \\
\hline
\texttt{LogicalGroup} &
First-class Boolean grouping of conditions. &
\texttt{operator\_type}, \texttt{evaluation\_order}, \texttt{expression\_tree}, \texttt{arity} &
$(A \wedge B) \vee (C \wedge D)$. \\
\hline
\texttt{ConditionGroup} &
Applicability, prerequisite, exception, or trigger condition group. &
\texttt{group\_type}, \texttt{evaluation}, \texttt{expression\_tree}, \texttt{description} &
All permit conditions met; emergency exception conditions. \\
\hline
\texttt{Defeasibility} &
Reified exception, override, or priority structure. &
\texttt{defeasibility\_type}, \texttt{description}, \texttt{rule\_scope}, \texttt{priority}, \texttt{condition} &
Special-event exception; emergency override. \\
\hline
\texttt{Formula} &
Mathematical, logical, or comparison expression &
\texttt{operation}, \texttt{operands}, \texttt{expression}, \texttt{comparator}, \texttt{result\_type}. &
temperature $< 41^\circ$F; distance $\leq$ 500 feet \\
\hline
\texttt{Metadata} &
Provenance, versioning, review, or annotation metadata. &
\texttt{source}, \texttt{comments}, \texttt{version}, \texttt{created\_by}, \texttt{review\_status}. &
Last updated 2024-01; manually validated. \\
\hline
\end{tabular}
\end{table*}

The most important normalization in Table~\ref{tab:olgpp_node_schema} is the treatment of \texttt{SemanticEntity}. We do not use it as an unconstrained catch-all node for every noun phrase. Instead, \texttt{SemanticEntity} is an abstract superclass for non-party domain entities that appear in legal rules. Concrete entities should be typed more specifically whenever possible. For example, \textit{raw chicken} may be represented as a food item, \textit{thermometer} as equipment, \textit{food truck} as a vehicle or facility, and \textit{sidewalk} as a location. This avoids using a single vague node type for parties, substances, equipment, locations, and legal artifacts.

\subsection{Normalized Edge Schema}

Table~\ref{tab:olgpp_edge_schema} gives the normalized OLG++ edge schema. Some edges are inherited from OLG, such as deontic modality, temporal relations, \texttt{WHAT\_REL}, and formula operations. Others are OLG++ additions required for regulatory settings. A few edges, such as \texttt{PERFORMED\_BY}, \texttt{HAS\_MEMBER}, \texttt{HAS\_JURISDICTION}, and \texttt{SUBCLASS\_OF}, are included because they are necessary to make the schema explicit and avoid leaving important relationships implicit in node labels.

\begin{table*}
\caption{Normalized OLG++ edge schema. All edges also have base properties \texttt{id}, \texttt{creation\_date}, \texttt{modified\_date}, and \texttt{temporal\_validity}.}
\label{tab:olgpp_edge_schema}
\centering
\small
\begin{tabular}{|p{3.0cm}|p{3.7cm}|p{3.9cm}|p{6.1cm}|}
\hline
\textbf{Category} & \textbf{Edge Label} & \textbf{Important Properties} & \textbf{Meaning and Example} \\
\hline

Core &
\texttt{DEONTIC\_MODALITY} &
\texttt{modality}, \texttt{strength}, \texttt{description} &
Links a trigger to the party to whom the modality applies. Modalities include obligation, prohibition, permission, and entitlement. \\
\hline

Core &
\texttt{WHAT\_REL} &
\texttt{description}, \texttt{validity} &
Links a trigger to the regulated action, object, or state of affairs. Example: ``may sell'' $\rightarrow$ ``sell food.'' \\
\hline

Core &
\texttt{PERFORMED\_BY} &
\texttt{role}, \texttt{description} &
Links an action or regulated activity to its actor. Example: ``sell food'' $\rightarrow$ food truck vendor. \\
\hline

Core &
\texttt{SUBJECT\_TO} &
\texttt{clause}, \texttt{jurisdiction}, \texttt{description} &
Links a trigger or action to a legal reference or governing condition. \\
\hline

Core &
\texttt{AMOUNT} &
\texttt{role}, \texttt{description} &
Links a rule, action, or formula to a numerical amount or threshold. \\
\hline

Logical Dependency &
\texttt{IF\_TRUE}, \texttt{IF\_FALSE}, \texttt{IF\_LATE} &
\texttt{condition}, \texttt{grace\_period}, \texttt{description} &
Links an event or condition to a consequent rule depending on truth, falsity, or lateness. \\
\hline

Logical Grouping &
\texttt{AND}, \texttt{OR}, \texttt{NOT}, \texttt{MEMBER} &
\texttt{evaluation\_order}, \texttt{position}, \texttt{role} &
Connects condition nodes to logical or condition groups. Example: two conjunctions connected by disjunction. \\
\hline

Temporal &
\texttt{BEFORE}, \texttt{AFTER}, \texttt{ON}, \texttt{DURING}, \texttt{RECURRING} &
\texttt{description}, \texttt{recurrence\_rule} &
Baseline temporal relations linking triggers, actions, or events to \texttt{Timex} nodes. \\
\hline

Temporal &
\texttt{TEMPORAL\_MODIFIER} &
\texttt{scope}, \texttt{description} &
Associates a rule or action with a temporal scope such as business hours or lunch hours. \\
\hline

Temporal &
\texttt{TEMPORAL\_OVERLAP}, \texttt{TEMPORAL\_SEQUENCE} &
\texttt{type}, \texttt{priority}, \texttt{gap}, \texttt{max\_gap} &
Represents concurrent or ordered obligations. Example: inspection within 30 minutes after a triggering event. \\
\hline

Location &
\texttt{SPATIAL} &
\texttt{spatial\_role}, \texttt{description} &
Links a rule or action to a location or jurisdictional context. \\
\hline

Location &
\texttt{LOCATION\_PREDICATE} &
\texttt{predicate\_type}, \texttt{parameters} &
Represents predicates such as within, outside, north of, or bounded by. \\
\hline

Location &
\texttt{PROXIMITY\_TO} &
\texttt{distance}, \texttt{unit}, \texttt{comparator} &
Represents distance constraints. Example: within 500 feet of a school. \\
\hline

Location &
\texttt{WITHIN}, \texttt{OUTSIDE}, \texttt{PART\_OF} &
\texttt{description} &
Represents spatial containment, exclusion, or part-whole structure. Example: Theatre District within Central Business District. \\
\hline

Jurisdiction &
\texttt{HAS\_JURISDICTION} &
\texttt{scope}, \texttt{validity} &
Links a jurisdiction to the rule that applies within it, or links a rule to its jurisdictional scope. \\
\hline

Party &
\texttt{HAS\_MEMBER}, \texttt{MEMBER\_OF}, \texttt{MEMBERSHIP} &
\texttt{role}, \texttt{period}, \texttt{rights} &
Represents party-group membership and collective participation. \\
\hline

Party Authority &
\texttt{DELEGATION}, \texttt{AGENCY} &
\texttt{type}, \texttt{duration}, \texttt{revocable}, \texttt{scope}, \texttt{limitations} &
Represents delegation or agency relationships among parties. \\
\hline

Conditional &
\texttt{PREREQUISITE}, \texttt{MUTUAL\_EXCLUSIVITY}, \texttt{CONTEXTUAL} &
\texttt{mandatory}, \texttt{grace\_period}, \texttt{scope}, \texttt{resolution}, \texttt{validity} &
Represents prerequisite requirements, mutually exclusive statuses, and contextual applicability conditions. \\
\hline

Defeasibility &
\texttt{EXCEPTION} &
\texttt{description}, \texttt{priority}, \texttt{condition}, \texttt{scope} &
Links a narrower exception rule or exception condition to the broader rule it defeats. \\
\hline

Defeasibility &
\texttt{OVERRIDE}, \texttt{PRECEDENCE} &
\texttt{precedence}, \texttt{level}, \texttt{basis}, \texttt{condition} &
Represents priority among rules when more than one rule applies. \\
\hline

Ontology Alignment &
\texttt{SUBCLASS\_OF}, \texttt{EQUIVALENT\_TO}, \texttt{RELATED\_TO} &
\texttt{ontology}, \texttt{confidence}, \texttt{validity}, \texttt{relation\_type} &
Links OLG++ entities to taxonomic or ontology-level relations. Example: sidewalk \texttt{SUBCLASS\_OF} public location. \\
\hline

Formula &
\texttt{FORMULA\_REL} and arithmetic/comparison edges &
\texttt{operation}, \texttt{operand\_role}, \texttt{position}, \texttt{operator}, \texttt{threshold} &
Connects formulas to operands and represents operations such as addition, maximum, or threshold comparison. \\
\hline

Extension Mechanism &
Declared domain-specific edge type &
\texttt{name}, \texttt{domain}, \texttt{range}, \texttt{properties}, \texttt{source\_ontology}, \texttt{justification} &
Allows OLG++ to reuse specialized relations from external ontologies or domain models, provided that each relation is explicitly declared with its domain, range, provenance, and modeling rationale; e.g., \texttt{STORED\_AT} from a food-safety ontology. \\
\hline

\end{tabular}
\end{table*}

The edge schema separates three notions that are often conflated in informal legal graphs. First, \texttt{SUBJECT\_TO} is a dependency on a legal reference or governing condition; it is not the same as a logical prerequisite. Second, \texttt{SPATIAL}, \texttt{PROXIMITY\_TO}, and \texttt{WITHIN} represent spatial semantics and should not be encoded merely as textual attributes on a \texttt{What} node. Third, \texttt{EXCEPTION}, \texttt{OVERRIDE}, and \texttt{PRECEDENCE} represent different forms of rule interaction and should not be collapsed into a single generic defeasibility edge.

\subsection{Semantic Entities and Ontology Alignment}

A central design choice in OLG++ is that domain entities should not be hidden inside long free-text node labels. In the baseline example from Section~\ref{sec:OLG}, the phrase ``public street, alley, highway, parking lot, sidewalk, or right-of-way'' should not be represented as one opaque \texttt{What} node. Instead, each location class should be represented as a separate \texttt{Location} or ontology-aligned \texttt{SemanticEntity} node, with a \texttt{SUBCLASS\_OF} edge to a more general concept such as \textit{public location}. The disjunction among these location classes should then be represented using a \texttt{LogicalGroup}.

This design does not require OLG++ to define every domain concept internally. If an external ontology already defines \textit{sidewalk}, \textit{street}, \textit{public right-of-way}, \textit{food truck}, or \textit{commercial kitchen}, the corresponding OLG++ node may store an \texttt{external\_ontology\_id} and may be connected using \texttt{SUBCLASS\_OF}, \texttt{EQUIVALENT\_TO}, or other alignment edges. OLG++ uses such domain entities only insofar as they participate in legal rules: as parties, regulated objects, locations, conditions, exceptions, or jurisdictional scopes.

\subsection{Locations and Jurisdictions}

Locations and jurisdictions are separated because they play different roles in regulatory language. A \texttt{Location} denotes a spatial entity or class of places, such as sidewalk, public park, school zone, or commercial kitchen. A \texttt{SpecificLocation} denotes a concrete named place or address. A \texttt{Jurisdiction} denotes the legal scope within which a rule applies, such as a city, county, business district, or cross-border regulatory authority.

For example, a rule may prohibit vending within 500 feet of a school. The school is a \texttt{Location} or \texttt{SpecificLocation}; the distance condition is represented by a \texttt{PROXIMITY\_TO} edge with \texttt{distance = 500} and \texttt{unit = feet}. By contrast, a rule applying only within the City of San Diego uses a \texttt{Jurisdiction} node connected to the rule by \texttt{HAS\_JURISDICTION}. If a theatre district is within a central business district, this is represented by a \texttt{WITHIN} edge between jurisdiction or location nodes. The distinction allows questions such as `Where is vending prohibited?'' and `Which jurisdiction's rules apply?'' to be answered by different graph patterns.

\subsection{Logical and Condition Groups}

OLG++ introduces \texttt{LogicalGroup} and \texttt{ConditionGroup} nodes to represent compound conditions explicitly. A \texttt{LogicalGroup} captures the Boolean structure of a condition, while a \texttt{ConditionGroup} captures the role of the condition in the legal rule, such as applicability, exception, prerequisite, or trigger condition.

Consider a food-truck rule stating that operation in a residential zone is allowed only if either (a) the operation occurs during daylight hours and is at least 500 feet from any school, or (b) the operation is part of a permitted neighborhood event and has written approval from 60
\[
(c_1 \wedge c_2) \vee (c_3 \wedge c_4),
\]
where $c_1$ is the daylight-hours condition, $c_2$ is the school-distance condition, $c_3$ is the neighborhood-event condition, and $c_4$ is the resident-approval condition. A \texttt{ConditionGroup} then links this expression to the rule as its applicability condition.

\subsection{Party Groups and Collective Obligations}

The \texttt{PartyGroup} node is used only when a legal provision refers to parties jointly, collectively, or by reference to a group. It is not a substitute for careful legal interpretation. OLG++ records the structural form of the group and the mode of responsibility, but it does not itself decide whether the law creates joint liability, several liability, joint-and-several liability, or merely coordinated duties. For this reason, \texttt{PartyGroup} includes a \texttt{liability\_mode} property.

For example, if a rule states that a food truck operator and a community kitchen manager must jointly sign and maintain a food safety inspection log, OLG++ represents the two parties as members of a \texttt{PartyGroup}. The obligation trigger is connected to the party group by a deontic modality edge. If, instead, the ordinance imposes separate obligations on the food truck operator and the kitchen manager, OLG++ represents them as distinct paths with separate obligation triggers. This distinction is important because a group representation should correspond to the legal structure of the provision, not merely to the fact that multiple parties are mentioned in the same sentence.

\subsection{Exceptions, Overrides, and Precedence}

OLG++ distinguishes exceptions, overrides, and precedence because they support different forms of legal explanation.

An \texttt{EXCEPTION} edge represents a rule-unless-condition structure. For example, ``food trucks may not vend in public parks unless they have a special event permit'' is represented as a general prohibition with an exception condition associated with the special event permit. The exception narrows or defeats the general rule under specified conditions.

An \texttt{OVERRIDE} edge represents a priority relation in which two rules may both apply, but one rule prevails. For example, a food truck may ordinarily be allowed to park in a permitted vending zone, but an emergency access rule may override that permission during a declared emergency. The ordinary permission is still a valid rule, but it loses priority in the emergency context.

A \texttt{PRECEDENCE} edge represents a more general ordering among rules, sources, or authorities. For example, state health-code requirements may have precedence over a local operating permission. OLG++ does not prescribe a single defeasible-logic proof theory. Instead, it makes the relevant rule-interaction structures explicit so that a reasoning system can apply the appropriate semantics for a particular legal or regulatory domain.

\subsection{Schema Integrity Constraints}

OLG++ imposes several modeling constraints intended to prevent the under-specification observed in the baseline model.

\begin{enumerate}[leftmargin=*]
\item A \texttt{What} node should represent one regulated action, object, or state of affairs. It should not contain an unparsed list of heterogeneous entities.
\item Enumerated domain concepts should be represented as separate nodes and connected by logical-group structure when the legal text uses conjunction or disjunction.
\item Locational expressions should be represented using \texttt{Location}, \texttt{SpecificLocation}, \texttt{Jurisdiction}, \texttt{LocationPredicate}, and spatial edges, rather than as text inside \texttt{What} nodes.
\item A \texttt{PartyGroup} should be used only when the legal provision creates or refers to a collective, joint, representative, or group-based party structure.
\item \texttt{EXCEPTION} and \texttt{OVERRIDE} should not be used interchangeably. Exceptions narrow applicability; overrides resolve priority when multiple rules apply.
\item Every node and edge extracted from text should retain source-span or source-reference information sufficient for validation and explanation.
\end{enumerate}

These constraints are part of the schema design. They are intended to make OLG++ graphs more queryable, explainable, and easier to validate than graphs whose nodes contain arbitrarily complex natural-language fragments.


\section{Using OLG++} 
\label{sec:using-olgpp} 

The previous section defined the normalized OLG++ schema. We now show how the schema is used in two complementary ways. First, we instantiate OLG++ graphs for representative regulatory provisions, showing how the schema captures deontic triggers, regulated actions, parties, temporal constraints, locations, jurisdictions, logical groupings, and defeasible rule interactions. Second, we show how these graph structures support query patterns for regulatory question answering. The purpose of this section is not to present a full empirical evaluation, but to demonstrate that the added structure in OLG++ is operationally useful: the constructs introduced in the schema correspond to graph patterns that can be traversed, filtered, and explained. 

\subsection{Worked OLG++ Representations} 
\label{sec:worked-examples}

This section illustrates how the normalized OLG++ schema represents regulatory provisions. The purpose is not to provide a complete extraction pipeline, but to show how the schema from Section~\ref{sec:OLG++} resolves the representational gaps identified in Section~\ref{sec:OLG}. We use examples from food-business and municipal regulation because these rules commonly combine deontic modality, regulated actions, parties, locations, temporal constraints, jurisdictional scope, and exceptions.

\subsubsection{A Simple Defeasible Vending Rule}

Consider the provision:

\begin{quote}
Food trucks are prohibited from vending within 500 feet of a school unless operating under a special event permit.
\end{quote}

This example contains a general prohibition, a regulated action, a party, a spatial constraint, a distance threshold, and an exception condition. In OLG++, these components are represented as explicit nodes and relationships rather than being collapsed into a single textual description. Table~\ref{tab:worked_school_nodes} shows the relevant nodes, and Table~\ref{tab:worked_school_edges} shows the edges.

\begin{table}[t]
\caption{OLG++ nodes for the school-proximity vending rule.}
\label{tab:worked_school_nodes}
\centering
\resizebox{\columnwidth}{!}{%
\begin{tabular}{|l|l|p{6.4cm}|}
\hline
\textbf{Node ID} & \textbf{Node Label} & \textbf{Meaning} \\
\hline
$o_1$ & \texttt{ObligationTrigger} &
General prohibition: food trucks are prohibited from vending near schools. \\
\hline
$p_1$ & \texttt{Party} &
Food truck operator or vendor. \\
\hline
$w_1$ & \texttt{What} &
Vending from a food truck. \\
\hline
$l_1$ & \texttt{Location} &
School. \\
\hline
$a_1$ & \texttt{Amount} &
500 feet. \\
\hline
$r_1$ & \texttt{Reference} &
Special event permit. \\
\hline
$c_1$ & \texttt{ConditionGroup} &
Condition that the food truck is operating under a valid special event permit. \\
\hline
$d_1$ & \texttt{Defeasibility} &
Exception structure for the special-event-permit case. \\
\hline
\end{tabular}
}
\end{table}

\begin{table}[t]
\caption{OLG++ edges for the school-proximity vending rule.}
\label{tab:worked_school_edges}
\centering
\resizebox{\columnwidth}{!}{%
\begin{tabular}{|l|l|l|p{5.8cm}|}
\hline
\textbf{Source} & \textbf{Edge Label} & \textbf{Target} & \textbf{Meaning} \\
\hline
$o_1$ & \texttt{DEONTIC\_MODALITY} & $p_1$ &
The prohibition applies to the food truck vendor. \\
\hline
$o_1$ & \texttt{WHAT\_REL} & $w_1$ &
The regulated action is vending. \\
\hline
$w_1$ & \texttt{PERFORMED\_BY} & $p_1$ &
The food truck vendor performs the vending action. \\
\hline
$w_1$ & \texttt{PROXIMITY\_TO} & $l_1$ &
The action is spatially constrained by proximity to a school. \\
\hline
$w_1$ & \texttt{AMOUNT} & $a_1$ &
The relevant distance threshold is 500 feet. \\
\hline
$d_1$ & \texttt{EXCEPTION} & $o_1$ &
The permit condition defeats the general prohibition when satisfied. \\
\hline
$d_1$ & \texttt{CONTEXTUAL} & $c_1$ &
The exception applies only in the special-event-permit context. \\
\hline
$c_1$ & \texttt{SUBJECT\_TO} & $r_1$ &
The condition is tied to the special event permit. \\
\hline
\end{tabular}
}
\end{table}

The example illustrates three design choices. First, the school is represented as a \texttt{Location}, not as text embedded inside the action node. Second, the distance threshold is represented as an \texttt{Amount}, allowing distance-based queries to be expressed over graph structure. Third, the permit is not represented merely as another conjunctive condition on the prohibition. It is represented through a \texttt{Defeasibility} node and an \texttt{EXCEPTION} edge, because the legal effect of the permit is to defeat the otherwise applicable prohibition.

\subsubsection{Representing Temporal and Locational Conditions}

Now consider the more complex provision introduced earlier:

\begin{quote}
A food truck vendor may not park or stand for longer than 60 minutes on any public street, alley, highway, parking lot, sidewalk, or right-of-way.
\end{quote}

The baseline OLG representation can identify the prohibition, party, action, and temporal expression. However, it tends to aggregate the list of public locations into a single node. OLG++ instead separates the temporal condition from the locational condition and represents the enumerated location classes explicitly. The prohibition applies when two primary conditions are both satisfied: the duration exceeds 60 minutes, and the location is one of the enumerated public-location types. The location alternatives are disjunctive, while the temporal and locational conditions are conjunctive.

Let $c_t$ denote the temporal condition that parking or standing exceeds 60 minutes. Let $c_l$ denote the locational condition that the activity occurs on a public street, alley, highway, parking lot, sidewalk, or right-of-way. The condition structure is $c_t \wedge c_l$, where $c_l$ expands to $\textit{street} \vee \textit{alley} \vee \textit{highway} \vee \textit{parking lot} \vee \textit{sidewalk} \vee \textit{right-of-way}$.

\begin{table*}
\caption{Selected OLG++ nodes for the 60-minute parking rule.}
\label{tab:worked_parking_nodes}
\centering
\small
\begin{tabular}{|l|l|p{10.2cm}|}
\hline
\textbf{Node ID} & \textbf{Node Label} & \textbf{Meaning} \\
\hline
$o_2$ & \texttt{ObligationTrigger} &
Prohibition trigger: a food truck vendor may not park or stand under the specified temporal and locational conditions. \\
\hline
$p_2$ & \texttt{Party} &
Food truck vendor. \\
\hline
$w_2$ & \texttt{What} &
Parking or standing a food truck. \\
\hline
$t_1$ & \texttt{Timex} &
Duration of 60 minutes. \\
\hline
$a_2$ & \texttt{Amount} &
Numerical threshold: 60 minutes. \\
\hline
$g_1$ & \texttt{LogicalGroup} &
Top-level conjunction: temporal condition and locational condition must both hold. \\
\hline
$g_2$ & \texttt{LogicalGroup} &
Location disjunction: one of the enumerated public-location classes must hold. \\
\hline
$l_2$--$l_7$ & \texttt{Location} &
Public street, alley, highway, parking lot, sidewalk, and right-of-way, represented as separate location nodes. \\
\hline
$l_8$ & \texttt{Location} &
General class: public location. \\
\hline
\end{tabular}
\end{table*}

\begin{table*}
\caption{Selected OLG++ edges for the 60-minute parking rule.}
\label{tab:worked_parking_edges}
\centering
\small
\begin{tabular}{|l|l|l|p{8.8cm}|}
\hline
\textbf{Source} & \textbf{Edge Label} & \textbf{Target} & \textbf{Meaning} \\
\hline
$o_2$ & \texttt{DEONTIC\_MODALITY} & $p_2$ &
The prohibition applies to the food truck vendor. \\
\hline
$o_2$ & \texttt{WHAT\_REL} & $w_2$ &
The regulated action is parking or standing. \\
\hline
$w_2$ & \texttt{PERFORMED\_BY} & $p_2$ &
The food truck vendor performs the regulated action. \\
\hline
$o_2$ & \texttt{CONTEXTUAL} & $g_1$ &
The prohibition is applicable under the top-level condition group. \\
\hline
$g_1$ & \texttt{AND} & $t_1$ &
The temporal condition is part of the top-level conjunction. \\
\hline
$g_1$ & \texttt{AND} & $g_2$ &
The locational disjunction is the second part of the top-level conjunction. \\
\hline
$t_1$ & \texttt{AMOUNT} & $a_2$ &
The temporal duration is tied to the numerical threshold of 60 minutes. \\
\hline
$g_2$ & \texttt{OR} & $l_2$--$l_7$ &
Each enumerated location class is an alternative sufficient to satisfy the locational condition. \\
\hline
$l_2$--$l_7$ & \texttt{SUBCLASS\_OF} & $l_8$ &
Each enumerated location class is represented as a subclass of public location. \\
\hline
$w_2$ & \texttt{SPATIAL} & $g_2$ &
The regulated action is spatially constrained by the locational condition group. \\
\hline
\end{tabular}
\end{table*}

This representation addresses two shortcomings of the baseline model. First, it prevents the list of locations from being hidden in one large \texttt{What} node. A query about sidewalks can now match the \texttt{Location} node for sidewalk and follow the \texttt{OR}, \texttt{SPATIAL}, and \texttt{WHAT\_REL} paths back to the relevant prohibition. Second, it records the correct logical structure: the location alternatives are disjunctive, while the temporal and locational requirements jointly determine applicability.

\subsubsection{Jurisdiction and Override}

The final example illustrates jurisdictional scope and rule priority:

\begin{quote}
Food trucks with vending permits may park in designated vending zones. During a declared emergency, emergency access rules override ordinary vending-zone permissions.
\end{quote}

This provision differs from the previous permit example. The declared emergency does not merely create an exception to one rule. Instead, it establishes a priority relation between two rules that may both be relevant: the ordinary vending-zone permission and the emergency access rule. OLG++ represents this situation using \texttt{OVERRIDE} or \texttt{PRECEDENCE}, rather than \texttt{EXCEPTION}.

\begin{table}[t]
\caption{Core OLG++ representation of jurisdiction and override.}
\label{tab:worked_override}
\centering
\resizebox{\columnwidth}{!}{%
\begin{tabular}{|l|l|p{6.4cm}|}
\hline
\textbf{Construct} & \textbf{OLG++ Element} & \textbf{Role} \\
\hline
Ordinary parking permission & \texttt{ObligationTrigger} $o_3$ &
Permission for permitted food trucks to park in designated vending zones. \\
\hline
Emergency access rule & \texttt{ObligationTrigger} $o_4$ &
Emergency rule restricting or redirecting parking during a declared emergency. \\
\hline
Declared emergency & \texttt{EventTrigger} $e_1$ &
Event that activates the emergency access rule. \\
\hline
Designated vending zone & \texttt{Location} $l_9$ &
Location class where ordinary vending-zone permission applies. \\
\hline
City jurisdiction & \texttt{Jurisdiction} $j_1$ &
Jurisdiction within which the vending-zone and emergency rules apply. \\
\hline
Rule priority & \texttt{OVERRIDE} edge $o_4 \rightarrow o_3$ &
The emergency rule takes priority over the ordinary permission when activated. \\
\hline
\end{tabular}
}
\end{table}

The important distinction is between exception and override. In the school-proximity example, the special event permit is an exception because it defeats a general prohibition under a specified permit condition. In the emergency example, the ordinary permission may remain valid as a rule, but the emergency rule has priority while the emergency event is active. The edge $o_4 \xrightarrow{\texttt{OVERRIDE}} o_3$ states that the emergency rule prevails over the ordinary vending-zone permission in the emergency context.

Jurisdiction is represented separately from ordinary location. The designated vending zone is a \texttt{Location}; the city or regulatory authority that defines the rule is a \texttt{Jurisdiction}. A \texttt{HAS\_JURISDICTION} edge connects the applicable rule to the jurisdictional scope. This allows a query to distinguish between where an activity occurs and which legal authority governs the activity.

\subsubsection{Summary of Representational Effects}

The three examples show how OLG++ changes the representation of regulatory provisions. First, regulated actions are separated from domain entities and locations. Second, locational and temporal constraints are represented as typed graph structure rather than natural-language fragments inside a node label. Third, nested logical conditions are represented by \texttt{LogicalGroup} and \texttt{ConditionGroup} nodes. Fourth, exceptions and overrides are distinguished as separate forms of defeasibility. Finally, jurisdictional scope is made explicit and queryable.

These modeling choices make the graph more verbose than baseline OLG, but the additional structure is precisely what allows regulatory questions to be answered by graph traversal and graph pattern matching rather than by reinterpreting node labels. The next section uses the worked representations above to illustrate case-study queries over OLG++ graphs.

\subsection{Querying OLG++ Representations} \label{sec:querying-olgpp} Once regulatory provisions are represented as OLG++ graphs, legal questions can be formulated as graph queries over deontic, spatial, temporal, jurisdictional, and defeasibility structure. We assume that OLG++ graphs are stored in a property graph database and write the examples in Cypher~\cite{francis2018cypher}. The queries below are intended as case-study query patterns rather than as a complete benchmark. They show how the graph representation supports questions that require combining multiple kinds of legal structure. A query over OLG++ typically follows five steps. First, it identifies the relevant regulated activity, represented by a \texttt{What} node. Second, it identifies the parties to whom the relevant deontic modality applies. Third, it constrains the query by temporal, spatial, jurisdictional, or amount conditions. Fourth, it checks whether exceptions, overrides, or precedence relations affect the rule. Finally, it returns the applicable rule, the legal source, and the graph path that explains why the rule applies.

\noindent\textbf{Query Pattern 1: Parking rules under temporal and spatial constraints.}
Consider the question:

\begin{quote}
What parking rules apply to food truck vendors in San Diego during lunch hours?
\end{quote}

This question is deliberately phrased as a rule-retrieval query rather than as a final legal-validity determination. Whether a particular parking location is ultimately valid may depend on the interaction of permissions, prohibitions, exceptions, overrides, and jurisdiction-specific rules. The role of the OLG++ query is therefore to retrieve the relevant deontic rules together with the contextual structure needed to interpret them.

The query in Figure~\ref{fig:cypher-parking-lunch} begins with the regulated action, represented by a \texttt{What} node whose \texttt{action\_type} is \texttt{parking}. It then follows the \texttt{PERFORMED\_BY} edge to restrict the actor to food truck vendors. Next, it retrieves all \texttt{ObligationTrigger} nodes connected to this parking action through \texttt{WHAT\_REL}, and filters them to rules whose \texttt{DEONTIC\_MODALITY} edge is either a permission or a prohibition. The query then restricts the result to rules within the San Diego jurisdiction, parking locations attached through \texttt{SPATIAL}, and temporal constraints that correspond to lunch hours. Finally, it collects any associated exceptions and overriding rules, so that each returned permission or prohibition is accompanied by the defeasible context needed for explanation.

\begin{figure*}[t]
\centering
\begin{minipage}{0.96\textwidth}
\small
\begin{lstlisting}[language=SQL,basicstyle=\ttfamily\small,breaklines=true,columns=fullflexible]
MATCH (w:What {action_type: 'parking'})
MATCH (w)-[:PERFORMED_BY]->(p:Party {party_type: 'food_truck_vendor'})

MATCH (o:ObligationTrigger)-[:WHAT_REL]->(w)
MATCH (o)-[dm:DEONTIC_MODALITY]->(p)
WHERE dm.modality IN ['permission', 'prohibition']

MATCH (j:Jurisdiction {name: 'San Diego'})
MATCH (o)-[:HAS_JURISDICTION]->(j)

MATCH (w)-[:SPATIAL]->(l:Location)
MATCH (o)-[:DURING|TEMPORAL_MODIFIER]->(t:Timex)
WHERE toLower(t.description) CONTAINS 'lunch'

OPTIONAL MATCH (d:Defeasibility)-[:EXCEPTION]->(o)
OPTIONAL MATCH (overridingRule:ObligationTrigger)-[:OVERRIDE]->(o)

WITH
  dm.modality AS Modality,
  w.description AS Activity,
  l.description AS Location,
  t.description AS Time,
  o.description AS Rule,
  collect(DISTINCT d.description) AS Exceptions,
  collect(DISTINCT overridingRule.description) AS OverridingRules

RETURN
  Modality,
  Activity,
  Location,
  Time,
  Rule,
  Exceptions,
  OverridingRules
ORDER BY
  Location,
  Modality;
\end{lstlisting}
\end{minipage}
\caption{Cypher query pattern for retrieving parking-related permissions and prohibitions for food truck vendors in San Diego during lunch hours. The query uses the OLG++ representation to combine the regulated action, party, jurisdiction, spatial scope, temporal scope, and defeasibility information. Exceptions and overriding rules are collected explicitly so that each returned rule can be interpreted with its defeasible context rather than as an isolated permission or prohibition.}
\label{fig:cypher-parking-lunch}
\end{figure*}

The important feature of this query is that the legal structure is represented graphically rather than textually. In the baseline OLG representation, parking locations, temporal scope, and exception information would either be absent or embedded in labels requiring additional text processing. In OLG++, the query can traverse explicit \texttt{Location}, \texttt{Jurisdiction}, \texttt{Timex}, \texttt{Defeasibility}, and \texttt{OVERRIDE} structures. The result is not merely a list of matching clauses; it is a structured explanation context showing which rule applies, where it applies, when it applies, and whether it is subject to exceptions or higher-priority rules.

\noindent\textbf{Query Pattern 2: Cross-jurisdictional food-safety obligations.}
Consider the question:

\begin{quote}
If a food truck prepares food in Tijuana but serves in San Diego, which safety regulations apply?
\end{quote}

This question illustrates a common regulatory pattern: the same business entity performs different regulated activities in different jurisdictions. Food preparation and food service are not merely two keywords in the same query. They are distinct regulated actions, each of which may be governed by a different legal authority, source document, prerequisite, or inspection regime. A representation that treats the food truck as a single undifferentiated business activity would miss this distinction. OLG++ instead represents \texttt{food\_preparation} and \texttt{food\_service} as separate \texttt{What} nodes, links each action to its spatial context, and then connects the relevant location to the jurisdiction in which that action occurs.

The query in Figure~\ref{fig:cypher-cross-jurisdiction} therefore has two branches. The first branch retrieves obligations governing food preparation when the preparation activity occurs within the Tijuana jurisdiction. The second branch retrieves obligations governing food service when the service activity occurs within the San Diego jurisdiction. Each branch follows the same OLG++ pattern: it starts from a regulated action, restricts that action to a jurisdiction through spatial containment, finds obligation triggers connected by \texttt{WHAT\_REL}, verifies that the obligation applies to the food-truck vendor through \texttt{DEONTIC\_MODALITY}, and then collects legal sources and prerequisite conditions. The final result is a jurisdiction-indexed set of obligations, sources, and prerequisites.

\begin{figure*}[t]
\centering
\begin{minipage}{0.96\textwidth}
\small
\begin{lstlisting}[language=SQL,basicstyle=\ttfamily\small,breaklines=true,columns=fullflexible]
MATCH (ft:Party {party_type: 'food_truck_vendor'})
MATCH (tj:Jurisdiction {name: 'Tijuana'})
MATCH (sd:Jurisdiction {name: 'San Diego'})
CALL {
  WITH ft, tj
  MATCH (prep:What {action_type: 'food_preparation'})
  MATCH (prep)-[:SPATIAL]->(prepLoc)
  MATCH (prepLoc)-[:WITHIN]->(tj)
  MATCH (op:ObligationTrigger)-[:WHAT_REL]->(prep)
  MATCH (op)-[:DEONTIC_MODALITY {modality: 'obligation'}]->(ft)
  OPTIONAL MATCH (op)-[:SUBJECT_TO]->(rp:Reference)
  OPTIONAL MATCH (op)-[:PREREQUISITE]->(preq1)
  RETURN
    'preparation' AS Activity,
    tj.name AS Jurisdiction,
    op.description AS Obligation,
    collect(DISTINCT rp.citation) AS Sources,
    collect(DISTINCT preq1.description) AS Prerequisites
}
RETURN
  Activity,
  Jurisdiction,
  Obligation,
  Sources,
  Prerequisites
UNION
MATCH (ft:Party {party_type: 'food_truck_vendor'})
MATCH (sd:Jurisdiction {name: 'San Diego'})
CALL {
  WITH ft, sd
  MATCH (serv:What {action_type: 'food_service'})
  MATCH (serv)-[:SPATIAL]->(servLoc)
  MATCH (servLoc)-[:WITHIN]->(sd)

  MATCH (os:ObligationTrigger)-[:WHAT_REL]->(serv)
  MATCH (os)-[:DEONTIC_MODALITY {modality: 'obligation'}]->(ft)

  OPTIONAL MATCH (os)-[:SUBJECT_TO]->(rs:Reference)
  OPTIONAL MATCH (os)-[:PREREQUISITE]->(preq2)
  RETURN
    'service' AS Activity,
    sd.name AS Jurisdiction,
    os.description AS Obligation,
    collect(DISTINCT rs.citation) AS Sources,
    collect(DISTINCT preq2.description) AS Prerequisites
}
RETURN
  Activity,
  Jurisdiction,
  Obligation,
  Sources,
  Prerequisites;
\end{lstlisting}
\end{minipage}
\caption{Cypher query pattern for retrieving food-safety obligations when food preparation and food service occur in different jurisdictions. It separately retrieves preparation obligations in Tijuana and service obligations in San Diego. Each branch preserves the OLG++ distinctions among regulated action, party, location, jurisdiction, legal source, and prerequisite conditions.}
\label{fig:cypher-cross-jurisdiction}
\end{figure*}

The key point is that the jurisdictional distinction is represented structurally. The query does not ask whether the word ``Tijuana'' or ``San Diego'' appears in the same clause as ``food safety.'' It follows explicit graph paths from regulated actions to spatial locations and from those locations to jurisdictions. This makes the returned answer explainable: each obligation is associated with the activity it governs, the jurisdiction in which that activity occurs, the legal source supporting the obligation, and any prerequisite conditions that must be satisfied. In the baseline OLG model, this distinction would be difficult to maintain because jurisdictional scope and spatial containment are not first-class constructs.

\noindent\textbf{What the Query Examples Demonstrate.} The query examples demonstrate that OLG++ is not merely a larger list of node and edge types. The added constructs correspond to graph paths needed for regulatory question answering. Location and jurisdiction nodes support spatial and multi-jurisdictional constraints. Logical and condition groups support nested applicability conditions. Defeasibility edges support exception and override checks. Reference nodes support source-level explanations. These examples do not constitute an empirical evaluation of legal question answering accuracy, nor do they solve the extraction problem. Rather, they show that once a provision is represented in OLG++, the graph contains the structure needed to formulate queries that would be difficult to express over the baseline OLG model without additional text processing or ad hoc interpretation of node labels.

\section{Preliminary Coverage Analysis}
\label{sec:evaluation}

We performed a first-pass coverage annotation over 35 provision units drawn from four chapters of the Carlsbad Municipal Code: Chapter 5.12 on cardroom regulation, Chapter 6.02 on county health and sanitation provisions including mobile food preparation units, Chapter 6.20 on single-use plastic foodware regulation, and Chapter 6.14 on smoking restrictions in unenclosed dining areas~\cite{carlsbad_cardrooms,carlsbad_county_health,carlsbad_foodware,carlsbad_smoking}. The goal was not to evaluate automatic extraction accuracy, but to test whether the normalized OLG++ schema provides sufficient representational coverage for real municipal regulatory provisions. The annotation was performed manually with LLM assistance and then organized according to the normalized node and edge types in Tables~\ref{tab:olgpp_node_schema} and~\ref{tab:olgpp_edge_schema}. The LLM was used as a drafting aid: given a provision unit and the OLG++ schema, it proposed candidate node and edge decompositions, which the authors then reviewed, corrected, normalized, and edited for consistency with the schema. We therefore do not treat the LLM output as an independent annotator, validator, or source of legal interpretation. The full first-pass annotations are provided in Appendix~\ref{app:olgpp-annotations}.

\subsection{Evaluation Setup}

The unit of analysis was a \emph{provision unit}: a definition, sentence, subsection, enumerated clause, or tightly connected group of clauses expressing one legal or semantic function. Each provision unit was annotated with OLG++ nodes, OLG++ edges, provision type, source section, and notes about modeling judgment. We deliberately selected provisions to exercise the full schema rather than to obtain a random sample of municipal law. This makes the study a \emph{coverage analysis}: it asks whether the schema can represent the variety of structures encountered in the selected provisions, not whether an automatic system can extract those structures unaided.

Table~\ref{tab:evaluation-corpus} summarizes the source chapters and the provision units selected from each. The sample includes constitutive definitions, prohibitions, permissions, permit requirements, exceptions, fee calculations, temporal rules, location restrictions, jurisdictional incorporation, enforcement provisions, and penalty provisions. This mixture was chosen because a useful regulatory representation must handle more than deontic modality alone.

\begin{table*}
\caption{Source chapters and provision units used in the preliminary coverage analysis.}
\label{tab:evaluation-corpus}
\centering
\small
\begin{tabular}{|p{2.6cm}|p{2.0cm}|p{4.8cm}|p{6.1cm}|}
\hline
\textbf{Source} & \textbf{Units} & \textbf{Provision Types} & \textbf{OLG++ Constructs Exercised} \\
\hline
Chapter 5.12 Cardrooms &
\texttt{C1--C14} &
Definitions, prohibitions, exceptions, work permits, operating rules, fee schedules, signage rules, penalties &
\texttt{SemanticEntity}, \texttt{Location}, \texttt{Party}, \texttt{PartyGroup}, \texttt{Amount}, \texttt{Timex}, \texttt{Formula}, \texttt{EXCEPTION}, \texttt{AGENCY}, \texttt{SUBCLASS\_OF}, \texttt{PART\_OF} \\
\hline
Chapter 6.02 County Code---Health and Sanitation &
\texttt{M1--M6} &
Incorporation by reference, mobile food facility grading, permit enforcement, health permit fees, mobile food preparation unit operation &
\texttt{Reference}, \texttt{Jurisdiction}, \texttt{EQUIVALENT\_TO}, \texttt{HAS\_JURISDICTION}, \texttt{PREREQUISITE}, \texttt{MULTIPLICATION}, \texttt{TEMPORAL\_SEQUENCE}, \texttt{TEMPORAL\_OVERLAP} \\
\hline
Chapter 6.20 Single-Use Plastic Foodware Ban &
\texttt{F1--F11} &
Definitions, food-service-provider taxonomy, compostability conditions, prohibitions, foodware-accessory request rules, exemptions, enforcement &
\texttt{ConditionGroup}, \texttt{LogicalGroup}, \texttt{SUBJECT\_TO}, \texttt{OVERRIDE}, \texttt{PRECEDENCE}, \texttt{EXCEPTION}, \texttt{DELEGATION}, \texttt{Formula}, \texttt{NOT}, \texttt{MUTUAL\_EXCLUSIVITY} \\
\hline
Chapter 6.14 Smoking in Unenclosed Dining Areas &
\texttt{S1--S4} &
Smoking prohibition, reasonable-distance restriction, state/federal law precedence, optional property-controller prohibition, signage and ashtray obligations &
\texttt{PROXIMITY\_TO}, \texttt{LocationPredicate}, \texttt{Amount}, \texttt{PRECEDENCE}, \texttt{HAS\_JURISDICTION}, \texttt{PART\_OF}, \texttt{NOT} \\
\hline
\end{tabular}
\end{table*}

\subsection{Coverage Results}

Table~\ref{tab:evaluation-summary} gives the aggregate results. The selected provisions collectively exercised every normalized OLG++ node type and every major edge category. Twenty-eight provision units were represented directly using the normalized schema. Seven provision units were also representable, but required modeling judgment involving exception/override distinctions, definition exclusions, party-group interpretation, jurisdictional scope, or incorporation by reference. No new primitive node or edge type was required. Three provisions used the declared extension mechanism for domain-specific relations: \texttt{USED\_FOR}, \texttt{ENTRANCE\_LEADS\_TO}, and \texttt{USED\_ALONGSIDE}. These were treated as declared domain-specific relations rather than as new OLG++ primitives because each can be specified with domain, range, provenance, and modeling justification.

\begin{table}[t]
\caption{First-pass coverage results over 35 provision units.}
\label{tab:evaluation-summary}
\centering
\small
\begin{tabular}{|p{5.4cm}|r|}
\hline
\textbf{Measure} & \textbf{Result} \\
\hline
Source chapters used & 4 \\
\hline
Provision units annotated & 35 \\
\hline
Units represented directly using normalized OLG++ & 28 \\
\hline
Units represented with modeling judgment & 7 \\
\hline
Units requiring a new primitive node type & 0 \\
\hline
Units requiring a new primitive edge type & 0 \\
\hline
Units using declared extension relations & 3 \\
\hline
Units involving deontic modality & 24 \\
\hline
Units involving definitions or constitutive rules & 8 \\
\hline
Units involving defeasibility, exception, override, or precedence & 12 \\
\hline
Units involving spatial, locational, or jurisdictional structure & 17 \\
\hline
Units involving temporal structure & 11 \\
\hline
Units involving formulas, amounts, thresholds, or fee calculations & 10 \\
\hline
Units involving external legal references or incorporation by reference & 12 \\
\hline
\end{tabular}
\end{table}

Table~\ref{tab:evaluation-construct-coverage} summarizes construct-level coverage. The cardroom provisions were especially useful for testing definitions, operating rules, numerical limits, recurring temporal closure, inspection obligations, fee formulas, and penalties. The mobile-food provisions in Chapter 6.02 tested incorporation by reference, jurisdictional substitution, mobile food facility enforcement, permit prerequisites, cost-recovery formulas, and operating restrictions. The foodware chapter tested rich definitions, exclusions, compostability conditions, food-service-provider taxonomies, foodware-accessory request rules, local/state precedence, exemptions, waivers, and enforcement. The smoking chapter tested proximity constraints, signage obligations, and state/federal precedence.

\begin{table*}
\caption{Coverage of major OLG++ construct families.}
\label{tab:evaluation-construct-coverage}
\centering
\small
\begin{tabular}{|p{3.7cm}|p{5.0cm}|p{7.0cm}|}
\hline
\textbf{Construct Family} & \textbf{Representative OLG++ Types} & \textbf{Examples from the Annotated Provisions} \\
\hline
Core deontic structure &
\texttt{ObligationTrigger}, \texttt{Party}, \texttt{What}, \texttt{DEONTIC\_MODALITY}, \texttt{WHAT\_REL}, \texttt{PERFORMED\_BY} &
Cardroom prohibition; work-permit obligation; foodware prohibition; mobile food preparation unit operating restrictions; smoking prohibition. \\
\hline
Definitions and ontology alignment &
\texttt{SemanticEntity}, \texttt{SUBCLASS\_OF}, \texttt{PART\_OF}, \texttt{EQUIVALENT\_TO}, \texttt{RELATED\_TO} &
Definition of cardroom; food-service-provider taxonomy; foodware and foodware-accessory definitions; county-to-city substitution in incorporated provisions. \\
\hline
Location and jurisdiction &
\texttt{Location}, \texttt{SpecificLocation}, \texttt{Jurisdiction}, \texttt{LocationPredicate}, \texttt{SPATIAL}, \texttt{WITHIN}, \texttt{PROXIMITY\_TO}, \texttt{HAS\_JURISDICTION} &
Carlsbad Senior Center exception; mobile food unit operation on public streets/private property; smoking within 20 feet of unenclosed dining areas; city facilities and city-affiliated events. \\
\hline
Logical and conditional structure &
\texttt{LogicalGroup}, \texttt{ConditionGroup}, \texttt{AND}, \texttt{OR}, \texttt{NOT}, \texttt{MEMBER}, \texttt{PREREQUISITE} &
Compostability conditions; recreational card-playing exception; foodware accessory request condition; mobile food seated-person condition; definition exclusions. \\
\hline
Temporal structure &
\texttt{Timex}, \texttt{BEFORE}, \texttt{AFTER}, \texttt{ON}, \texttt{DURING}, \texttt{RECURRING}, \texttt{TEMPORAL\_SEQUENCE}, \texttt{TEMPORAL\_OVERLAP} &
Appeal within 10 days; one-year work permit; daily 2:00 a.m.--10:00 a.m. cardroom closure; educational-material posting within 30 days; payment within 15 days; cooking while operating a mobile food unit. \\
\hline
Amounts and formulas &
\texttt{Amount}, \texttt{Formula}, \texttt{AMOUNT}, \texttt{FORMULA\_REL}, arithmetic and comparison edges &
Cardroom table limits and player limits; work-permit fee; cardroom fee schedule; penalties up to \$1,000; enforcement costs up to three times permit cost; daily foodware penalties capped annually. \\
\hline
Defeasibility and priority &
\texttt{Defeasibility}, \texttt{EXCEPTION}, \texttt{OVERRIDE}, \texttt{PRECEDENCE}, \texttt{CONTEXTUAL} &
Recreational card-playing exception; Senior Center exception; foil-wrapper and straw exceptions; emergency/no-feasible-alternative foodware exemptions; local chapter governing when more stringent than state law; state/federal smoking-law precedence. \\
\hline
Authority and group structure &
\texttt{PartyGroup}, \texttt{HAS\_MEMBER}, \texttt{MEMBER\_OF}, \texttt{DELEGATION}, \texttt{AGENCY} &
Chief of Police work-permit authority; City Manager exemption authority; city departments, agents, employees, and designees; operator/employee supervision of card tables. \\
\hline
References and provenance &
\texttt{Reference}, \texttt{Metadata}, \texttt{SUBJECT\_TO} &
County code incorporated by reference; California Retail Food Code permit reference; AB 1276 reference; Penal Code reference; source-section metadata for every annotation unit. \\
\hline
\end{tabular}
\end{table*}

\subsection{Modeling Judgments}

The annotation revealed several recurring modeling judgments. These are not failures of coverage; rather, they identify where annotation guidelines and human validation are needed. Table~\ref{tab:evaluation-modeling-judgments} lists the principal cases.

\begin{table*}
\caption{Recurring modeling judgments in the first-pass annotations.}
\label{tab:evaluation-modeling-judgments}
\centering
\small
\begin{tabular}{|p{3.0cm}|p{4.7cm}|p{7.7cm}|}
\hline
\textbf{Issue} & \textbf{Example Provisions} & \textbf{Modeling Resolution} \\
\hline
Exception vs. override vs. precedence &
Recreational card-playing exception; Senior Center exception; local foodware chapter more stringent than AB 1276; state/federal smoking-law scope &
Use \texttt{EXCEPTION} when a narrower condition defeats a general rule; use \texttt{OVERRIDE} or \texttt{PRECEDENCE} when two rules may both apply but one has priority. \\
\hline
Definition exclusions &
Foodware definition excluding egg cartons, meat trays, coolers, ice chests, and packing materials; food-service-provider exclusions &
Represent included classes with \texttt{SUBCLASS\_OF}; represent exclusions with \texttt{NOT}, \texttt{MUTUAL\_EXCLUSIVITY}, or a set-difference \texttt{Formula} when the definition is explicitly subtractive. \\
\hline
Party groups and responsibility mode &
City departments, agents, employees, and designees; cardroom operator and employees &
Use \texttt{PartyGroup} only when the text creates a collective or role-based group. Record responsibility using properties such as \texttt{role}, \texttt{liability\_mode}, or \texttt{agency\_scope}; do not infer liability beyond the text. \\
\hline
Location vs. jurisdiction &
City facilities, public streets, private property, Carlsbad Senior Center, City of Carlsbad &
Use \texttt{Location} or \texttt{SpecificLocation} for places where activities occur; use \texttt{Jurisdiction} for the legal authority or territorial scope of applicability. \\
\hline
Incorporation by reference &
County health and sanitation provisions adopted into the city code; county-to-city substitution rule &
Use \texttt{Reference} and \texttt{SUBJECT\_TO} for incorporated law; use \texttt{EQUIVALENT\_TO} or substitution mappings for county terms mapped to city counterparts; use \texttt{AGENCY} where the county acts on behalf of the city. \\
\hline
Domain-specific relations &
Card table used for card games; cardroom entrance leading to liquor-serving establishment; accessories used alongside prepared food &
Do not introduce new primitive OLG++ edge types. Use the declared extension mechanism and require each relation to specify name, domain, range, source, and modeling justification. \\
\hline
Formula normalization &
Cardroom fee schedule; enforcement costs up to three times permit cost; daily penalty capped annually &
Represent arithmetic structure using \texttt{Formula}, \texttt{FORMULA\_REL}, and arithmetic/comparison edges, while preserving the source text for validation. \\
\hline
\end{tabular}
\end{table*}

\subsection{Interpretation of the Results}

The coverage analysis supports three observations. First, the selected provisions exercised the full range of constructs introduced by OLG++. The examples required not only deontic triggers and parties, but also definitions, domain entities, locations, jurisdictions, logical groups, temporal constraints, formulae, references, exceptions, overrides, and authority relations. This supports the design decision to treat OLG++ as a richer regulatory representation layer rather than as a minor syntactic extension of OLG.

Second, the added structure was needed for queryability and explanation. For example, cardroom rules require recurring temporal closure and police inspection during operating hours; mobile-food provisions require incorporation by reference and permit-related enforcement formulas; foodware provisions require definition exclusions, request-based conditions, local/state precedence, and temporary exemptions; smoking provisions require proximity constraints and state/federal precedence. These structures would be difficult to query reliably if represented only as long labels on \texttt{what} nodes.

Third, the analysis identifies the boundary between schema coverage and legal interpretation. OLG++ can represent candidate structures for exceptions, overrides, party groups, and jurisdictional scope, but the legally correct interpretation may require human validation. For that reason, the appendix annotations should be read as first-pass schema-coverage annotations, not as authoritative legal interpretations. The result is nevertheless useful: for the sampled provisions, the normalized schema covered the representational needs without requiring new primitive node or edge types, while also exposing the annotation guidelines needed for future corpus-scale evaluation.

\subsection{Threats to Validity}

This preliminary analysis has several limitations. The provisions were selected to exercise the schema and are not a random sample of municipal law. The annotations were produced as first-pass LLM-assisted annotations and were not independently validated by legal experts or multiple annotators. The counts therefore measure representational coverage under a controlled annotation exercise, not extraction accuracy, legal correctness, or inter-annotator agreement. In addition, some provision units contain multiple legal effects, so alternative segmentation choices could change the unit counts.

Despite these limitations, the study provides a useful stress test of OLG++ on real municipal provisions. It shows that the schema can cover a diverse set of regulatory constructs, and it makes explicit where future work should focus: annotation guidelines, gold-standard datasets, inter-annotator agreement, automated extraction, ontology alignment, and formal semantics for exceptions, overrides, and precedence.

\section{Discussion and Conclusion}
\label{sec:discussion-conclusion}

The examples in Section~\ref{sec:using-olgpp} show how OLG++ can represent and query regulatory provisions whose meaning depends on combinations of deontic modality, regulated action, party role, location, jurisdiction, temporal scope, and defeasible rule interaction. This section clarifies the scope of the model, its relationship to other legal and semantic representation frameworks, and the limitations that remain.

\noindent\textbf{OLG++ as a representation layer.}
OLG++ is best understood as a representation layer for regulatory rules, not as a complete ontology or a complete legal reasoning system. Its purpose is to make explicit the graph structure needed to connect legal rules to the entities, actions, events, locations, times, quantities, and conditions that determine their applicability. In this sense, OLG++ extends OLG by preserving its compact deontic core while adding typed constructs needed for municipal and regulatory settings.

This distinction is important because many concepts that occur in regulatory text are not themselves deontic concepts. Terms such as \textit{school}, \textit{sidewalk}, \textit{food truck}, \textit{thermometer}, \textit{raw chicken}, and \textit{commercial kitchen} are domain concepts. OLG++ does not attempt to define all such concepts internally. Instead, it provides places in the graph where these concepts can be attached to obligations, permissions, prohibitions, exceptions, and jurisdictional scopes. When external vocabularies or ontologies are available, OLG++ nodes can be aligned to them using properties such as \texttt{external\_ontology\_id} and edges such as \texttt{SUBCLASS\_OF}, \texttt{EQUIVALENT\_TO}, and \texttt{RELATED\_TO}. Thus, OLG++ should be viewed as complementary to foundational, domain, and geospatial ontologies rather than as a replacement for them.

\noindent\textbf{Relationship to LegalRuleML.}
LegalRuleML provides a standard XML-based language for representing legal rules, including rule statements, defeasibility, temporal qualifications, and references to legal sources~\cite{legalruleml-core-1.0,athan2015legalruleml}. OLG++ has a different design goal. Rather than serving primarily as an interchange syntax for legal rules, it is designed as a property-graph representation that supports graph traversal, graph querying, and explanation paths over regulatory provisions. This makes OLG++ closer to a graph-oriented knowledge representation layer for legal question answering.

The two approaches are therefore not mutually exclusive. A LegalRuleML representation could supply a formal rule-oriented view of a provision, while OLG++ could provide a graph-oriented view that connects the same provision to locations, parties, regulated actions, jurisdictional scopes, source clauses, exceptions, and external ontologies. For example, an exception structure in LegalRuleML may correspond in OLG++ to an \texttt{EXCEPTION} edge from a defeasibility node or exception rule to the defeated general rule. Similarly, a rule-priority relation may correspond to an \texttt{OVERRIDE} or \texttt{PRECEDENCE} edge. OLG++ makes these relationships available as graph patterns that can be queried and inspected directly.

At the same time, OLG++ does not by itself prescribe a complete proof theory for defeasible legal reasoning. The model distinguishes exception, override, and precedence relationships because these distinctions are important for explanation and querying. However, the interpretation of these relationships may depend on the intended legal reasoning semantics, such as defeasible logic, an argumentation framework, or a domain-specific compliance engine. In this paper, we focus on representing these structures explicitly rather than committing to one reasoning calculus.

\noindent\textbf{Extraction and validation.}
This paper focuses on the representation problem, but an operational OLG++ system also requires an extraction and validation pipeline. Legal provisions must be segmented into clauses; definitions, parties, actions, temporal expressions, quantities, locations, and references must be identified; and relations such as \texttt{WHAT\_REL}, \texttt{PERFORMED\_BY}, \texttt{SPATIAL}, \texttt{HAS\_JURISDICTION}, and \texttt{EXCEPTION} must be extracted or validated. Some of these tasks can be supported by NLP methods, dependency parsing, named-entity recognition, semantic role labeling, and large language models. Others require domain-specific normalization, ontology alignment, and human review.

OLG++ is designed with this extraction setting in mind. Nodes and edges can carry source-span information, confidence scores, provenance, review status, and links to legal sources. These properties are important because regulatory question answering should not merely return an answer; it should also explain which provision, clause, definition, or exception supports the answer. A graph node or edge without provenance is difficult to validate and difficult to use in a high-stakes compliance setting. Future work should therefore treat extraction, normalization, and human validation as first-class components of an OLG++ construction pipeline.

\noindent\textbf{Expressivity and modeling tradeoffs.}
The added expressivity of OLG++ comes with modeling tradeoffs. A more explicit graph is larger than a baseline OLG representation. A provision that could be represented as one trigger node and one large \texttt{What} node in OLG may become a graph involving a regulated action, several location nodes, an amount node, a temporal expression, a logical group, and one or more ontology-alignment edges. This increased size is justified when the goal is queryability and explanation. For example, representing \textit{street}, \textit{sidewalk}, \textit{parking lot}, and \textit{right-of-way} as separate location nodes makes it possible to answer sidewalk-specific questions by graph traversal rather than by string matching inside a node label.

There is also a boundary question: which relations should be native to OLG++ and which should be imported from external ontologies? Our position is that OLG++ should include relations that are necessary to connect legal rules to their applicability structure, such as \texttt{WHAT\_REL}, \texttt{DEONTIC\_MODALITY}, \texttt{HAS\_JURISDICTION}, \texttt{EXCEPTION}, and \texttt{OVERRIDE}. More specialized domain relations, such as food-storage relations or detailed geospatial topology, may be imported from domain or geospatial ontologies where appropriate. OLG++ therefore provides a normalized legal-regulatory graph backbone, while allowing domain-specific extensions.

\noindent\textbf{Limitations and future work.}
The current work has three main limitations. First, the paper presents a schema and worked examples, but not a full empirical evaluation over a corpus of regulations. The Cypher queries in Section~\ref{sec:using-olgpp} are case-study queries intended to demonstrate the operational use of the representation, not a benchmark of question-answering accuracy. Second, the paper does not provide an automatic extraction algorithm from legal text to OLG++ graphs. Such an algorithm would need to combine linguistic analysis, ontology alignment, and validation. Third, although OLG++ represents defeasible structures such as exceptions and overrides, it does not yet define a full formal semantics for computing legal conclusions under conflicts, priorities, and exceptions ~\cite{prakken2015law}.

These limitations define a research agenda. A complete OLG++ system should include a graph-construction pipeline, a validation workflow, a library of reusable regulatory modeling patterns, and a reasoning layer that can interpret the graph under a chosen legal semantics. It should also be evaluated on a corpus of municipal and regulatory provisions to measure coverage, extraction difficulty, graph quality, and usefulness for legal question answering.

In conclusion, OLG++ extends the Obligation Logic Graph from a compact representation of deontic content to a normalized property-graph schema for regulatory and municipal legal knowledge. Its central contribution is to make explicit the structures that are often implicit in legal text: regulated actions, parties, locations, jurisdictions, temporal constraints, logical groupings, exceptions, overrides, precedence relations, and links to legal sources. By making these structures first-class graph elements, OLG++ supports more transparent graph queries and more explainable regulatory question answering. The model is not a substitute for legal interpretation, external ontologies, or defeasible reasoning engines; rather, it provides a semantic graph substrate on which such components can operate.

\bibliographystyle{ACM-Reference-Format}
\bibliography{icail}


\begin{thebibliography}{11}


\ifx \showCODEN    \undefined \def \showCODEN     #1{\unskip}     \fi
\ifx \showDOI      \undefined \def \showDOI       #1{#1}\fi
\ifx \showISBNx    \undefined \def \showISBNx     #1{\unskip}     \fi
\ifx \showISBNxiii \undefined \def \showISBNxiii  #1{\unskip}     \fi
\ifx \showISSN     \undefined \def \showISSN      #1{\unskip}     \fi
\ifx \showLCCN     \undefined \def \showLCCN      #1{\unskip}     \fi
\ifx \shownote     \undefined \def \shownote      #1{#1}          \fi
\ifx \showarticletitle \undefined \def \showarticletitle #1{#1}   \fi
\ifx \showURL      \undefined \def \showURL       {\relax}        \fi
\providecommand\bibfield[2]{#2}
\providecommand\bibinfo[2]{#2}
\providecommand\natexlab[1]{#1}
\providecommand\showeprint[2][]{arXiv:#2}

\bibitem[Angles(2018)]%
        {angles2018property}
\bibfield{author}{\bibinfo{person}{Renzo Angles}.} \bibinfo{year}{2018}\natexlab{}.
\newblock \showarticletitle{The Property Graph Database Model.}. In \bibinfo{booktitle}{\emph{Proc. of the AMW}}.
\newblock


\bibitem[Athan et~al\mbox{.}(2015)]%
        {athan2015legalruleml}
\bibfield{author}{\bibinfo{person}{Tara Athan}, \bibinfo{person}{Guido Governatori}, \bibinfo{person}{Monica Palmirani}, \bibinfo{person}{Adrian Paschke}, {and} \bibinfo{person}{Adam Wyner}.} \bibinfo{year}{2015}\natexlab{}.
\newblock \showarticletitle{LegalRuleML: Design principles and foundations}.
\newblock \bibinfo{journal}{\emph{Reasoning Web. Web Logic Rules: 11th International Summer School 2015, Berlin, Germany, July 31-August 4, 2015, Tutorial Lectures. 11}} (\bibinfo{year}{2015}), \bibinfo{pages}{151--188}.
\newblock


\bibitem[Boer et~al\mbox{.}(2008)]%
        {boer2008metalex}
\bibfield{author}{\bibinfo{person}{Alexander Boer}, \bibinfo{person}{Radboud Winkels}, {and} \bibinfo{person}{Fabio Vitali}.} \bibinfo{year}{2008}\natexlab{}.
\newblock \showarticletitle{Metalex XML and the legal knowledge interchange format}.
\newblock In \bibinfo{booktitle}{\emph{Computable Models of the Law: Languages, Dialogues, Games, Ontologies}}. \bibinfo{publisher}{Springer}, \bibinfo{pages}{21--41}.
\newblock


\bibitem[Filtz(2021)]%
        {filtz2021knowledge}
\bibfield{author}{\bibinfo{person}{Erwin Filtz}.} \bibinfo{year}{2021}\natexlab{}.
\newblock \emph{\bibinfo{title}{Knowledge Graphs for Analyzing and Searching Legal Data}}.
\newblock \bibinfo{thesistype}{Master's\ thesis}. \bibinfo{school}{Inst. of Data, Process and Knowl. Management, Vienna University of Economics and Business}.
\newblock


\bibitem[Francis et~al\mbox{.}(2018)]%
        {francis2018cypher}
\bibfield{author}{\bibinfo{person}{Nadime Francis}, \bibinfo{person}{Alastair Green}, \bibinfo{person}{Paolo Guagliardo}, \bibinfo{person}{Leonid Libkin}, \bibinfo{person}{Tobias Lindaaker}, \bibinfo{person}{Victor Marsault}, \bibinfo{person}{Stefan Plantikow}, \bibinfo{person}{Mats Rydberg}, \bibinfo{person}{Petra Selmer}, {and} \bibinfo{person}{Andr{\'e}s Taylor}.} \bibinfo{year}{2018}\natexlab{}.
\newblock \showarticletitle{Cypher: An evolving query language for property graphs}. In \bibinfo{booktitle}{\emph{Proc. of the 2018 Int. Conf. on management of data}}. \bibinfo{pages}{1433--1445}.
\newblock


\bibitem[Hoekstra et~al\mbox{.}(2007)]%
        {hoekstra2007lkif}
\bibfield{author}{\bibinfo{person}{Rinke Hoekstra}, \bibinfo{person}{Joost Breuker}, \bibinfo{person}{Marcello~Di Bello}, {and} \bibinfo{person}{Alexander Boer}.} \bibinfo{year}{2007}\natexlab{}.
\newblock \showarticletitle{{The LKIF Core Ontology of Basic Legal Concepts}}. In \bibinfo{booktitle}{\emph{Proceedings of the Workshop on Legal Ontologies and Artificial Intelligence Techniques}} \emph{(\bibinfo{series}{CEUR Workshop Proceedings}, Vol.~\bibinfo{volume}{321})}. \bibinfo{pages}{43--63}.
\newblock


\bibitem[Martinez-Gil(2023)]%
        {martinez2023survey}
\bibfield{author}{\bibinfo{person}{Jorge Martinez-Gil}.} \bibinfo{year}{2023}\natexlab{}.
\newblock \showarticletitle{A survey on legal question--answering systems}.
\newblock \bibinfo{journal}{\emph{Computer Science Review}}  \bibinfo{volume}{48} (\bibinfo{year}{2023}), \bibinfo{pages}{100552}.
\newblock


\bibitem[Palmirani et~al\mbox{.}(2021)]%
        {legalruleml-core-1.0}
\bibfield{author}{\bibinfo{person}{Monica Palmirani}, \bibinfo{person}{Guido Governatori}, \bibinfo{person}{Tara Athan}, \bibinfo{person}{Harold Boley}, \bibinfo{person}{Adrian Paschke}, {and} \bibinfo{person}{Adam Wyner}.} \bibinfo{year}{2021}\natexlab{}.
\newblock \bibinfo{title}{{LegalRuleML Core Specification Version 1.0}}.
\newblock \bibinfo{howpublished}{OASIS Standard}.
\newblock
\urldef\tempurl%
\url{https://docs.oasis-open.org/legalruleml/legalruleml-core-spec/v1.0/legalruleml-core-spec-v1.0.html}
\showURL{%
\tempurl}
\newblock
\shownote{Edited by Monica Palmirani, Guido Governatori, Tara Athan, Harold Boley, Adrian Paschke, and Adam Wyner}.


\bibitem[Prakken and Sartor(2015)]%
        {prakken2015law}
\bibfield{author}{\bibinfo{person}{Henry Prakken} {and} \bibinfo{person}{Giovanni Sartor}.} \bibinfo{year}{2015}\natexlab{}.
\newblock \showarticletitle{Law and Logic: A Review from an Argumentation Perspective}.
\newblock \bibinfo{journal}{\emph{Artificial Intelligence}}  \bibinfo{volume}{227} (\bibinfo{year}{2015}), \bibinfo{pages}{214--245}.
\newblock
\urldef\tempurl%
\url{https://doi.org/10.1016/j.artint.2015.06.005}
\showDOI{\tempurl}


\bibitem[Schneider et~al\mbox{.}(2022)]%
        {schneider2022lynx}
\bibfield{author}{\bibinfo{person}{Juli{\'a}n~Moreno Schneider}, \bibinfo{person}{Georg Rehm}, \bibinfo{person}{Elena Montiel-Ponsoda}, \bibinfo{person}{V{\'\i}ctor Rodr{\'\i}guez-Doncel}, \bibinfo{person}{Patricia Mart{\'\i}n-Chozas}, \bibinfo{person}{Mar{\'\i}a Navas-Loro}, \bibinfo{person}{Martin Kaltenb{\"o}ck}, \bibinfo{person}{Artem Revenko}, \bibinfo{person}{Sotirios Karampatakis}, \bibinfo{person}{Christian Sageder}, {et~al\mbox{.}}} \bibinfo{year}{2022}\natexlab{}.
\newblock \showarticletitle{{Lynx: A Knowledge-Based AI Service Platform for Content Processing, Enrichment and Analysis for the Legal Domain}}.
\newblock \bibinfo{journal}{\emph{Information Systems}}  \bibinfo{volume}{106} (\bibinfo{year}{2022}), \bibinfo{pages}{101966}.
\newblock
\urldef\tempurl%
\url{https://doi.org/10.1016/j.is.2021.101966}
\showDOI{\tempurl}


\bibitem[Servantez et~al\mbox{.}(2023)]%
        {servantez2023computable}
\bibfield{author}{\bibinfo{person}{Sergio Servantez}, \bibinfo{person}{Nedim Lipka}, \bibinfo{person}{Alexa Siu}, \bibinfo{person}{Milan Aggarwal}, \bibinfo{person}{Balaji Krishnamurthy}, \bibinfo{person}{Aparna Garimella}, \bibinfo{person}{Kristian Hammond}, {and} \bibinfo{person}{Rajiv Jain}.} \bibinfo{year}{2023}\natexlab{}.
\newblock \bibinfo{title}{Computable contracts by extracting obligation logic graphs}.
\newblock , \bibinfo{numpages}{267--276}~pages.
\newblock


\end{thebibliography}

\clearpage
\appendix
\section{First-Pass OLG++ Annotations for Carlsbad Regulatory Provisions}
\label{app:olgpp-annotations}

This appendix gives first-pass OLG++ annotations for 35 provision units drawn from selected chapters of the Carlsbad Municipal Code. The goal is schema-coverage analysis rather than legal advice or gold-standard annotation. Each provision unit is represented using the notation below.

\noindent\textbf{Notation.} A node is written as:\\ \texttt{nodeId:NodeType(label; properties)}.\\
An edge is written as:\\
\texttt{(source)-[:EDGE\_TYPE \{properties\}]->(target)}. \\
Property lists are illustrative and focus on legally relevant fields. Every node is assumed to have the base OLG++ node properties \texttt{id}, \texttt{created\_date}, \texttt{modified\_date}, and \texttt{status}. Every edge is assumed to have the base OLG++ edge properties \texttt{id}, \texttt{creation\_date}, \texttt{modified\_date}, and \texttt{temporal\_validity}. Each annotation should also carry \texttt{source\_ref} and \texttt{source\_text} provenance in an implementation.

\lstdefinestyle{olgppann}{
  basicstyle=\ttfamily\scriptsize,
  breaklines=true,
  breakatwhitespace=false,
  columns=fullflexible,
  keepspaces=true,
  linewidth=\columnwidth,
  xleftmargin=0pt,
  xrightmargin=0pt,
  aboveskip=0.4em,
  belowskip=0.4em,
  frame=single,
  framerule=0.2pt
}
\lstnewenvironment{olgppgraph}{\lstset{style=olgppann}}{}

Table \ref{tab:declared-domain-relations} presents the domain-specific relationships used in the OLG++ representation.

\begin{table*}[t]
\caption{Declared domain-specific relations used by the first-pass annotations. These are not new primitive OLG++ edge types; they are extension-mechanism relations declared with domain, range, and modeling justification.}
\label{tab:declared-domain-relations}
\centering
\small
\begin{tabular}{|p{2.8cm}|p{3.0cm}|p{3.0cm}|p{6.2cm}|}
\hline
\textbf{Relation} & \textbf{Domain} & \textbf{Range} & \textbf{Justification} \\
\hline
\texttt{USED\_FOR} & \texttt{SemanticEntity}, \texttt{Location}, or \texttt{What} & \texttt{What} or \texttt{SemanticEntity} & Used for constitutive definitions where an entity is defined by intended use; e.g., a card table used or intended to be used for playing cards. \\
\hline
\texttt{ENTRANCE\_LEADS\_TO} & \texttt{Location} & \texttt{Location} & Used for physical access/connection relation in the cardroom alcohol provision. This is more specific than generic \texttt{SPATIAL} or \texttt{PROXIMITY\_TO}. \\
\hline
\texttt{USED\_ALONGSIDE} & \texttt{SemanticEntity} & \texttt{SemanticEntity} or \texttt{What} & Used for foodware accessory definitions where items are used as part of or alongside prepared food. \\
\hline
\end{tabular}
\end{table*}

\subsection{Cardroom Regulations: Chapter 5.12}

\subsubsection*{C1. Section 5.12.010: Definition of Cardroom}
\noindent\textbf{Provision unit.} A cardroom is any space, room, collection of rooms, or enclosure furnished or equipped with a table used or intended to be used as a card table for playing cards and similar games.

\noindent\textbf{Nodes.}
\begin{olgppgraph}
C1_ref:Reference(Section 5.12.010; source=Carlsbad Municipal Code, clause=Definition)
C1_cardroom:SemanticEntity(cardroom; semantic_type=defined_term)
C1_space:Location(space/room/collection/enclosure; location_type=indoor_space)
C1_table:SemanticEntity(card table; semantic_type=equipment)
C1_play:What(playing cards and similar games; action_type=game_play)
\end{olgppgraph}

\noindent\textbf{Edges.}
\begin{olgppgraph}
(C1_cardroom)-[:SUBJECT_TO]->(C1_ref)
(C1_cardroom)-[:SUBCLASS_OF]->(C1_space)
(C1_table)-[:PART_OF]->(C1_cardroom)
(C1_table)-[:USED_FOR {declared=true}]->(C1_play)
(C1_cardroom)-[:CONTEXTUAL {role=definition}]->(C1_play)
\end{olgppgraph}

\noindent\textbf{Annotation note.} This is a constitutive definition rather than a deontic rule. It exercises \texttt{SemanticEntity}, \texttt{Location}, \texttt{PART\_OF}, \texttt{SUBCLASS\_OF}, and a declared use relation.

\subsubsection*{C2. Section 5.12.020: Cardrooms Prohibited}
\noindent\textbf{Provision unit.} It is unlawful for any person to engage in, carry on, maintain, conduct, or cause a cardroom in the city for playing cards or games of chance where cards are used for money, profit, barter, or other consideration.

\noindent\textbf{Nodes.}
\begin{olgppgraph}
C2_ref:Reference(Section 5.12.020; clause=Cardrooms prohibited)
C2_o:ObligationTrigger(cardroom prohibition; modality=prohibition)
C2_person:Party(any person; party_type=person)
C2_act:What(engage/carry on/maintain/conduct/cause cardroom; action_type=operate_cardroom)
C2_cardroom:SemanticEntity(cardroom; semantic_type=defined_term)
C2_city:Jurisdiction(City of Carlsbad; jurisdiction_type=city)
C2_purpose:ConditionGroup(cards or games of chance involving consideration; group_type=purpose_condition)
C2_consideration:LogicalGroup(money/profit/barter/other consideration; operator_type=OR)
\end{olgppgraph}

\noindent\textbf{Edges.}
\begin{olgppgraph}
(C2_o)-[:SUBJECT_TO]->(C2_ref)
(C2_o)-[:DEONTIC_MODALITY {modality=prohibition}]->(C2_person)
(C2_o)-[:WHAT_REL]->(C2_act)
(C2_act)-[:PERFORMED_BY]->(C2_person)
(C2_act)-[:RELATED_TO {role=regulated_object}]->(C2_cardroom)
(C2_o)-[:HAS_JURISDICTION]->(C2_city)
(C2_o)-[:CONTEXTUAL]->(C2_purpose)
(C2_purpose)-[:MEMBER]->(C2_consideration)
(C2_consideration)-[:OR]->(money/profit/barter/other_consideration)
\end{olgppgraph}

\noindent\textbf{Annotation note.} The action alternatives and consideration alternatives should be represented as explicit logical alternatives rather than embedded in a single text label.

\subsubsection*{C3. Section 5.12.020: Recreational Playing in Single-Family Homes}
\noindent\textbf{Provision unit.} Recreational playing of cards in single-family homes is not intended to be prohibited where no consideration passes to the owner.

\noindent\textbf{Nodes.}
\begin{olgppgraph}
C3_d:Defeasibility(recreational home exception; defeasibility_type=exception)
C3_play:What(recreational card playing; action_type=recreational_play)
C3_home:Location(single-family home; location_type=residential)
C3_owner:Party(owner of residence; party_type=property_owner)
C3_cond:ConditionGroup(no consideration passes to owner; group_type=exception_condition)
C3_not:LogicalGroup(no consideration; operator_type=NOT)
\end{olgppgraph}

\noindent\textbf{Edges.}
\begin{olgppgraph}
(C3_d)-[:EXCEPTION]->(C2_o)
(C3_d)-[:CONTEXTUAL]->(C3_cond)
(C3_play)-[:SPATIAL]->(C3_home)
(C3_cond)-[:MEMBER]->(C3_not)
(C3_not)-[:NOT]->(consideration_to_owner)
(consideration_to_owner)-[:PERFORMED_BY]->(C3_owner)
\end{olgppgraph}

\noindent\textbf{Annotation note.} The phrase ``not intended to prohibit'' is treated as an exception to the general cardroom prohibition, not as an independent permission.

\subsubsection*{C4. Section 5.12.025: Senior Center Exception}
\noindent\textbf{Provision unit.} Chapter 5.12 does not apply to lawful playing of card games at the Carlsbad Senior Center at times and under reasonable rules and regulations established by the City Manager.

\noindent\textbf{Nodes.}
\begin{olgppgraph}
C4_d:Defeasibility(Senior Center exception; defeasibility_type=exception)
C4_play:What(lawful playing of card games; action_type=card_game_play)
C4_loc:SpecificLocation(Carlsbad Senior Center)
C4_time:Timex(times established by City Manager; temporal_type=authorized_time)
C4_rules:ConditionGroup(reasonable rules and regulations; group_type=authority_condition)
C4_manager:Party(City Manager; party_type=public_official)
\end{olgppgraph}

\noindent\textbf{Edges.}
\begin{olgppgraph}
(C4_d)-[:EXCEPTION]->(Chapter_5_12)
(C4_d)-[:CONTEXTUAL]->(C4_rules)
(C4_play)-[:SPATIAL]->(C4_loc)
(C4_play)-[:DURING]->(C4_time)
(C4_manager)-[:AGENCY {scope=establish_rules}]->(C4_rules)
\end{olgppgraph}

\noindent\textbf{Annotation note.} This is a location-specific chapter-level exception conditioned by authorized times and rules.

\subsubsection*{C5. Section 5.12.030: Existing Lawful Use Excepted}
\noindent\textbf{Provision unit.} Cardrooms lawfully existing on the date of passage may continue, provided they are not altered, improved, reconstructed, restored, repaired, intensified, expanded, or extended.

\noindent\textbf{Nodes.}
\begin{olgppgraph}
C5_perm:ObligationTrigger(existing cardroom continuation; modality=permission)
C5_event:EventTrigger(existing on passage date; event_type=lawful_preexisting_use)
C5_date:Timex(date of passage; temporal_type=date)
C5_cardroom:SemanticEntity(existing cardroom)
C5_cond:ConditionGroup(no prohibited modification; group_type=continuation_condition)
C5_or:LogicalGroup(alter/improve/reconstruct/restore/repair/intensify/expand/extend; operator_type=OR)
C5_not:LogicalGroup(not modified; operator_type=NOT)
\end{olgppgraph}

\noindent\textbf{Edges.}
\begin{olgppgraph}
(C5_event)-[:ON]->(C5_date)
(C5_event)-[:IF_TRUE]->(C5_perm)
(C5_perm)-[:WHAT_REL]->(continue_existing_cardroom)
(continue_existing_cardroom)-[:RELATED_TO]->(C5_cardroom)
(C5_perm)-[:CONTEXTUAL]->(C5_cond)
(C5_cond)-[:MEMBER]->(C5_not)
(C5_not)-[:NOT]->(C5_or)
(C5_or)-[:OR]->(alter/improve/reconstruct/restore/repair/intensify/expand/extend)
\end{olgppgraph}

\noindent\textbf{Annotation note.} This is a grandfathering permission with a negated disjunction of prohibited modifications.

\subsubsection*{C6. Section 5.12.040: Work Permit Requirements}
\noindent\textbf{Provision unit.} Cardroom employees must obtain work permits from the Chief of Police; permit applications are submitted under oath; a \$10 fee accompanies the application; issued permits are valid for one year.

\noindent\textbf{Nodes.}
\begin{olgppgraph}
C6_o:ObligationTrigger(obtain work permit; modality=obligation)
C6_employee:Party(cardroom employee; role=employee)
C6_chief:Party(Chief of Police; role=issuing_authority)
C6_permit:Reference(cardroom work permit)
C6_application:What(submit application under oath; action_type=application)
C6_fee:Amount(\$10.00; value=10, currency=USD, quantity_type=fee)
C6_validity:Timex(one year; temporal_type=duration)
\end{olgppgraph}

\noindent\textbf{Edges.}
\begin{olgppgraph}
(C6_o)-[:DEONTIC_MODALITY {modality=obligation}]->(C6_employee)
(C6_o)-[:WHAT_REL]->(C6_application)
(C6_application)-[:SUBJECT_TO]->(C6_permit)
(C6_employee)-[:PREREQUISITE]->(C6_permit)
(C6_application)-[:AMOUNT {role=application_fee}]->(C6_fee)
(C6_permit)-[:DURING {role=validity}]->(C6_validity)
(C6_chief)-[:AGENCY {scope=issue_or_deny_permit}]->(C6_permit)
\end{olgppgraph}

\noindent\textbf{Annotation note.} This provision combines a permit prerequisite, issuing authority, fee, and temporal validity.

\subsubsection*{C7. Section 5.12.050: Revocation/Suspension and Appeal}
\noindent\textbf{Provision unit.} The Chief of Police may revoke or suspend a cardroom license or work permit for cause; the action is subject to appeal to the City Council; notice of appeal must be filed within 10 days or the action becomes final and conclusive.

\noindent\textbf{Nodes.}
\begin{olgppgraph}
C7_event:EventTrigger(revocation or suspension; event_type=administrative_action)
C7_chief:Party(Chief of Police; role=revoking_authority)
C7_license:Reference(cardroom license or work permit)
C7_appeal:ObligationTrigger(appeal entitlement/procedure; modality=entitlement)
C7_council:Party(City Council; role=appeal_authority)
C7_collector:Party(license collector; role=filing_recipient)
C7_ten:Timex(10 days; temporal_type=deadline)
C7_final:EventTrigger(action becomes final; event_type=finality)
\end{olgppgraph}

\noindent\textbf{Edges.}
\begin{olgppgraph}
(C7_chief)-[:AGENCY {scope=revoke_or_suspend}]->(C7_event)
(C7_event)-[:SUBJECT_TO]->(C7_license)
(C7_event)-[:IF_TRUE]->(C7_appeal)
(C7_appeal)-[:DEONTIC_MODALITY {modality=entitlement}]->(licensee)
(C7_appeal)-[:SUBJECT_TO]->(C7_council)
(file_appeal)-[:BEFORE]->(C7_ten)
(file_appeal)-[:PERFORMED_BY]->(licensee)
(file_appeal)-[:SUBJECT_TO]->(C7_collector)
(C7_appeal)-[:IF_LATE]->(C7_final)
\end{olgppgraph}

\noindent\textbf{Annotation note.} This provision exercises \texttt{IF\_LATE}, temporal deadline semantics, administrative authority, and appeal entitlement.

\subsubsection*{C8. Section 5.12.060: License Number, Transfer, and Automatic Revocation}
\noindent\textbf{Provision unit.} No person may hold more than one cardroom license; licenses are not assignable or transferable; a license is automatically revoked if operation does not begin within six months, subject to one additional six-month extension by written application to the City Council.

\noindent\textbf{Nodes.}
\begin{olgppgraph}
C8_limit:ObligationTrigger(one license limit; modality=prohibition)
C8_person:Party(person/licensee)
C8_license:Reference(cardroom license)
C8_one:Amount(one license; value=1, quantity_type=count)
C8_nontransfer:ObligationTrigger(nonassignable/nontransferable license; modality=prohibition)
C8_event:EventTrigger(no operation begun; event_type=failure_to_commence)
C8_six:Timex(six months; temporal_type=duration)
C8_revocation:EventTrigger(automatic revocation; event_type=revocation)
C8_extension:Defeasibility(additional six-month extension; defeasibility_type=exception)
\end{olgppgraph}

\noindent\textbf{Edges.}
\begin{olgppgraph}
(C8_limit)-[:DEONTIC_MODALITY {modality=prohibition}]->(C8_person)
(C8_limit)-[:AMOUNT {role=max_licenses}]->(C8_one)
(C8_person)-[:MUTUAL_EXCLUSIVITY {scope=license_count}]->(more_than_one_license)
(C8_nontransfer)-[:WHAT_REL]->(assign_or_transfer_license)
(C8_event)-[:AFTER]->(C8_six)
(C8_event)-[:IF_TRUE]->(C8_revocation)
(C8_extension)-[:EXCEPTION]->(C8_revocation)
(written_application)-[:SUBJECT_TO]->(City_Council)
\end{olgppgraph}

\noindent\textbf{Annotation note.} The extension clause is modeled as an exception to automatic revocation, not as a general permission to delay.

\subsubsection*{C9. Section 5.12.070(A--G): Cardroom Operating Rules}
\noindent\textbf{Provision unit.} Cardroom operation is unlawful if it violates enumerated rules including one cardroom per address, permitted games only, no more than five tables, no more than eight players per table, no minors at tables or games, closure from 2:00 a.m. to 10:00 a.m. every day, and police inspection during operating hours.

\noindent\textbf{Nodes.}
\begin{olgppgraph}
C9_o:ObligationTrigger(operate in violation of rules; modality=prohibition)
C9_cardroom:SemanticEntity(cardroom)
C9_addr:Location(address)
C9_one:Amount(one cardroom; value=1)
C9_games:LogicalGroup(permitted games; operator_type=OR)
C9_tables:Amount(five tables; value=5)
C9_players:Amount(eight players; value=8)
C9_minor:Party(minor; party_type=minor)
C9_closed:Timex(2:00 a.m. to 10:00 a.m. every day; recurrence=daily)
C9_police:Party(police; party_type=enforcement_agency)
C9_inspection:What(police inspection; action_type=inspection)
\end{olgppgraph}

\noindent\textbf{Edges.}
\begin{olgppgraph}
(C9_o)-[:WHAT_REL]->(operate_cardroom)
(C9_cardroom)-[:SPATIAL]->(C9_addr)
(C9_cardroom)-[:AMOUNT {role=max_per_address}]->(C9_one)
(game_played)-[:MEMBER]->(C9_games)
(C9_games)-[:OR]->(pinochle/low_ball/draw_poker/panguingue/bridge)
(card_tables)-[:AMOUNT {role=max_tables}]->(C9_tables)
(players_per_table)-[:AMOUNT {role=max_players}]->(C9_players)
(minor_participation)-[:DEONTIC_MODALITY {modality=prohibition}]->(C9_minor)
(C9_cardroom)-[:DURING {state=closed}]->(C9_closed)
(C9_closed)-[:RECURRING {frequency=daily}]->(every_day)
(C9_inspection)-[:PERFORMED_BY]->(C9_police)
(C9_inspection)-[:DURING]->(hours_of_operation)
(C9_inspection)-[:TEMPORAL_OVERLAP]->(hours_of_operation)
\end{olgppgraph}

\noindent\textbf{Annotation note.} This single section exercises cardinality, allowed-enumeration, minor exclusion, recurring temporal closure, and inspection access.

\subsubsection*{C10. Section 5.12.080: Alcoholic Beverages Prohibited}
\noindent\textbf{Provision unit.} Alcoholic liquor or beverage may not be served, consumed, sold, or given away in any cardroom; no cardroom may have an entrance leading to an establishment that serves or sells intoxicating liquor.

\noindent\textbf{Nodes.}
\begin{olgppgraph}
C10_o:ObligationTrigger(alcohol prohibition; modality=prohibition)
C10_alcohol:SemanticEntity(alcoholic liquor or beverage)
C10_actions:LogicalGroup(serve/consume/sell/give away; operator_type=OR)
C10_cardroom:Location(cardroom)
C10_est:Location(liquor-serving establishment)
C10_entrance:LocationPredicate(entrance leading to establishment)
\end{olgppgraph}

\noindent\textbf{Edges.}
\begin{olgppgraph}
(C10_o)-[:WHAT_REL]->(C10_actions)
(C10_actions)-[:OR]->(serve/consume/sell/give_away)
(serve/consume/sell/give_away)-[:RELATED_TO]->(C10_alcohol)
(serve/consume/sell/give_away)-[:SPATIAL]->(C10_cardroom)
(C10_cardroom)-[:ENTRANCE_LEADS_TO {declared=true}]->(C10_est)
(C10_cardroom)-[:LOCATION_PREDICATE]->(C10_entrance)
\end{olgppgraph}

\noindent\textbf{Annotation note.} \texttt{ENTRANCE\_LEADS\_TO} is declared as a domain-specific spatial/access relation.

\subsubsection*{C11. Sections 5.12.090--5.12.150: Rates, Stakes, and Monthly Collections}
\noindent\textbf{Provision unit.} The code sets maximum participation charges, maximum bets/table stakes, monthly table fees, and revenue-percentage license fees.

\noindent\textbf{Nodes.}
\begin{olgppgraph}
C11_rate:Formula(game participation rates; operation=rate_schedule)
C11_stakes:Formula(table stakes limit; operation=comparison)
C11_monthly:Formula(monthly collections; operation=fee_schedule)
C11_amounts:Amount(20 cents, 15 cents, 5%, 60 cents, $20, $100, $30, $5, 7%, 8%, 9%)
C11_quarter:Timex(payable quarterly in advance)
C11_revenue:Amount(monthly gross revenue thresholds)
\end{olgppgraph}

\noindent\textbf{Edges.}
\begin{olgppgraph}
(C11_rate)-[:FORMULA_REL]->(C11_amounts)
(poker_fee)-[:MULTIPLICATION]->(5_percent_of_pot)
(bridge_fee)-[:DIVISION {basis=per_hour_per_player}]->(60_cents)
(C11_stakes)-[:COMPARISON {operator=less_than_or_equal}]->($20_bet/$100_hand)
(C11_monthly)-[:ADDITION]->(base_table_fee + revenue_percentage_fee)
(revenue_percentage_fee)-[:MULTIPLICATION]->(monthly_gross_revenue)
(C11_monthly)-[:RECURRING {frequency=monthly}]->(month)
(payment)-[:DURING]->(C11_quarter)
\end{olgppgraph}

\noindent\textbf{Annotation note.} This cluster is used primarily to exercise \texttt{Formula}, arithmetic edges, comparison, recurring fee cycles, and thresholds.

\subsubsection*{C12. Section 5.12.120: Supervision and Identification Badges}
\noindent\textbf{Provision unit.} Card tables must be supervised by the operator or employees; they may refuse participation; they must ensure operation complies with the chapter and state Penal Code; operators and employees must wear identification badges while on duty.

\noindent\textbf{Nodes.}
\begin{olgppgraph}
C12_group:PartyGroup(operator or employees; group_type=operational_staff)
C12_operator:Party(operator)
C12_employee:Party(employee)
C12_supervise:What(supervise card tables)
C12_refuse:ObligationTrigger(refuse participation; modality=permission)
C12_badge:What(wear identification badge)
C12_onduty:Timex(when on duty)
C12_penal:Reference(State Penal Code)
\end{olgppgraph}

\noindent\textbf{Edges.}
\begin{olgppgraph}
(C12_group)-[:HAS_MEMBER]->(C12_operator)
(C12_group)-[:HAS_MEMBER]->(C12_employee)
(C12_operator)-[:MEMBER_OF]->(C12_group)
(C12_employee)-[:MEMBER_OF]->(C12_group)
(C12_supervise)-[:PERFORMED_BY]->(C12_group)
(C12_refuse)-[:DEONTIC_MODALITY {modality=permission}]->(C12_group)
(operate_cardroom)-[:SUBJECT_TO]->(Chapter_5_12)
(operate_cardroom)-[:SUBJECT_TO]->(C12_penal)
(C12_badge)-[:PERFORMED_BY]->(C12_group)
(C12_badge)-[:DURING]->(C12_onduty)
\end{olgppgraph}

\noindent\textbf{Annotation note.} This provision uses \texttt{PartyGroup} because the same operational rule applies to operator or employees.

\subsubsection*{C13. Sections 5.12.130--5.12.140: Exterior and Interior Signs}
\noindent\textbf{Provision unit.} Exterior signs advertising cardrooms are prohibited except specified signs, with size limits; interior signs must state permitted games and charges and be visible from all parts of the cardroom.

\noindent\textbf{Nodes.}
\begin{olgppgraph}
C13_ext:ObligationTrigger(exterior sign restriction; modality=prohibition)
C13_int:ObligationTrigger(interior sign posting; modality=obligation)
C13_sign:SemanticEntity(sign)
C13_premises:Location(cardroom premises)
C13_interior:Location(interior of cardroom)
C13_size:Amount(1.5 feet by 6 feet; quantity_type=dimension)
C13_content:SemanticEntity(sign content: permitted games and charges)
\end{olgppgraph}

\noindent\textbf{Edges.}
\begin{olgppgraph}
(C13_ext)-[:WHAT_REL]->(advertising_sign)
(advertising_sign)-[:SPATIAL]->(C13_premises)
(C13_sign)-[:AMOUNT {role=max_size}]->(C13_size)
(C13_size)-[:COMPARISON {operator=less_than_or_equal}]->(allowed_dimension)
(C13_int)-[:WHAT_REL]->(post_interior_signs)
(post_interior_signs)-[:SPATIAL]->(C13_interior)
(C13_content)-[:PART_OF]->(C13_sign)
\end{olgppgraph}

\noindent\textbf{Annotation note.} This provision exercises sign-content, visibility/location, and amount/dimension constraints.

\subsubsection*{C14. Section 5.12.170: Violations}
\noindent\textbf{Provision unit.} Violation or failure to comply with Chapter 5.12 is unlawful, constitutes a misdemeanor, and is punishable by fine up to \$1,000, imprisonment up to six months, or both.

\noindent\textbf{Nodes.}
\begin{olgppgraph}
C14_event:EventTrigger(violation/failure to comply; event_type=violation)
C14_penalty:ObligationTrigger(misdemeanor penalty; modality=sanction)
C14_person:Party(violating person)
C14_fine:Amount(\$1,000 maximum; value=1000, currency=USD)
C14_jail:Amount(six months maximum; value=6, unit=months)
C14_choice:LogicalGroup(fine or imprisonment or both; operator_type=OR/AND)
\end{olgppgraph}

\noindent\textbf{Edges.}
\begin{olgppgraph}
(C14_event)-[:IF_TRUE]->(C14_penalty)
(C14_penalty)-[:DEONTIC_MODALITY {modality=sanction}]->(C14_person)
(C14_penalty)-[:AMOUNT]->(C14_fine)
(C14_penalty)-[:AMOUNT]->(C14_jail)
(C14_fine)-[:MAXIMUM]->(\$1000)
(C14_jail)-[:MAXIMUM]->(six_months)
(C14_choice)-[:OR]->(fine)
(C14_choice)-[:OR]->(imprisonment)
(C14_choice)-[:AND]->(fine_and_imprisonment)
\end{olgppgraph}

\noindent\textbf{Annotation note.} Penalty alternatives exercise both disjunction and combined sanction semantics.

\subsection*{Mobile Food and County Health Code Provisions: Chapter 6.02}

\subsubsection*{M1. Section 6.02.010(B): Mobile Food Facility Grading System Adopted by Reference}
\noindent\textbf{Provision unit.} A San Diego County mobile-food-facility grading system is adopted by reference and incorporated into the city code; county references are interpreted as corresponding city entities, or the county acts on behalf of the city when no city counterpart exists.

\noindent\textbf{Nodes.}
\begin{olgppgraph}
M1_citycode:Reference(Carlsbad Code)
M1_countycode:Reference(San Diego County Code Title 6, Division 1, Chapter 1, Section 61.101)
M1_grading:SemanticEntity(mobile food facility grading system)
M1_city:Jurisdiction(City of Carlsbad)
M1_county:Jurisdiction(County of San Diego)
M1_cityentity:Party(corresponding city board/agency/official/employee)
M1_countyentity:Party(county board/agency/official/employee)
\end{olgppgraph}

\noindent\textbf{Edges.}
\begin{olgppgraph}
(M1_citycode)-[:SUBJECT_TO {role=incorporates}]->(M1_countycode)
(M1_grading)-[:SUBJECT_TO]->(M1_countycode)
(M1_grading)-[:HAS_JURISDICTION]->(M1_city)
(M1_countyentity)-[:EQUIVALENT_TO {context=city_interpretation}]->(M1_cityentity)
(M1_countyentity)-[:AGENCY {scope=acts_on_behalf_of_city, condition=no_city_counterpart}]->(M1_city)
(M1_countycode)-[:HAS_JURISDICTION]->(M1_county)
\end{olgppgraph}

\noindent\textbf{Annotation note.} This provision is a strong example of incorporation by reference and jurisdictional substitution.

\subsubsection*{M2. Section 6.02.020: Mobile Food Facility Without Required Permit}
\noindent\textbf{Provision unit.} If DEH initiates enforcement against a person operating a mobile food facility without a CRFC-required permit, DEH may recover enforcement costs up to three times the cost of the permit.

\noindent\textbf{Nodes.}
\begin{olgppgraph}
M2_event:EventTrigger(operating mobile food facility without required permit)
M2_deh:Party(County DEH; role=enforcement_agency)
M2_person:Party(violator/person operating facility)
M2_facility:SemanticEntity(mobile food facility)
M2_permit:Reference(CRFC-required permit)
M2_recover:ObligationTrigger(recover enforcement costs; modality=permission)
M2_formula:Formula(up to three times permit cost; operation=multiplication_with_maximum)
M2_three:Amount(three times; value=3, unit=multiplier)
\end{olgppgraph}

\noindent\textbf{Edges.}
\begin{olgppgraph}
(M2_event)-[:RELATED_TO]->(M2_facility)
(M2_event)-[:PREREQUISITE {negated=true}]->(M2_permit)
(M2_event)-[:IF_TRUE]->(M2_recover)
(M2_recover)-[:DEONTIC_MODALITY {modality=permission}]->(M2_deh)
(M2_recover)-[:AMOUNT]->(M2_formula)
(M2_formula)-[:MULTIPLICATION]->(M2_three)
(M2_formula)-[:FORMULA_REL {operand=permit_cost}]->(M2_permit)
(M2_formula)-[:MAXIMUM]->(three_times_permit_cost)
(M2_deh)-[:AGENCY]->(enforcement_action)
\end{olgppgraph}

\noindent\textbf{Annotation note.} This provision exercises permit prerequisite failure, enforcement event, agency, multiplication, and maximum cap.

\subsubsection*{M3. Section 6.02.020: Payment Timing for Assessment}
\noindent\textbf{Provision unit.} After enforcement is completed, DEH may send an assessment; the violator must pay within 15 days from assessment or when applying for the permit, whichever occurs first.

\noindent\textbf{Nodes.}
\begin{olgppgraph}
M3_complete:EventTrigger(enforcement completed)
M3_assessment:EventTrigger(penalty assessment sent)
M3_pay:ObligationTrigger(pay assessment; modality=obligation)
M3_violator:Party(violator)
M3_fifteen:Timex(15 days from assessment)
M3_permitapp:EventTrigger(violator applies for permit)
M3_earlier:Formula(earlier of 15 days or permit application; operation=minimum)
\end{olgppgraph}

\noindent\textbf{Edges.}
\begin{olgppgraph}
(M3_complete)-[:AFTER]->(enforcement_activity)
(M3_complete)-[:IF_TRUE]->(M3_assessment)
(M3_assessment)-[:IF_TRUE]->(M3_pay)
(M3_pay)-[:DEONTIC_MODALITY {modality=obligation}]->(M3_violator)
(M3_pay)-[:TEMPORAL_SEQUENCE {max_gap=15_days}]->(M3_fifteen)
(M3_earlier)-[:MINIMUM]->(M3_fifteen)
(M3_earlier)-[:MINIMUM]->(M3_permitapp)
(M3_pay)-[:BEFORE]->(M3_earlier)
\end{olgppgraph}

\noindent\textbf{Annotation note.} The phrase ``whichever occurs first'' is modeled as a \texttt{MINIMUM} temporal formula.

\subsubsection*{M4. Section 6.02.030: Health Permit Fees}
\noindent\textbf{Provision unit.} All persons and businesses required to obtain a health-related permit or service from DEH must pay the county-established fee, including delinquent payment fees.

\noindent\textbf{Nodes.}
\begin{olgppgraph}
M4_pay:ObligationTrigger(pay health permit/service fee; modality=obligation)
M4_party:Party(person or business)
M4_deh:Party(DEH)
M4_permit:Reference(health-related permit or service)
M4_fee:Amount(county-established fee)
M4_latefee:Amount(delinquent payment fee)
\end{olgppgraph}

\noindent\textbf{Edges.}
\begin{olgppgraph}
(M4_pay)-[:DEONTIC_MODALITY {modality=obligation}]->(M4_party)
(M4_party)-[:PREREQUISITE]->(M4_permit)
(M4_permit)-[:SUBJECT_TO]->(M4_deh)
(M4_pay)-[:AMOUNT {role=fee}]->(M4_fee)
(M4_pay)-[:AMOUNT {role=delinquent_fee}]->(M4_latefee)
\end{olgppgraph}

\noindent\textbf{Annotation note.} This is a standard fee obligation triggered by permit/service requirement.

\subsubsection*{M5. Section 6.02.040(A): Seated-Person Condition for Mobile Food Preparation Units}
\noindent\textbf{Provision unit.} No person may drive or operate a mobile food preparation unit on public street or private property unless all persons in the vehicle are seated.

\noindent\textbf{Nodes.}
\begin{olgppgraph}
M5_o:ObligationTrigger(drive/operate prohibition unless seated; modality=prohibition)
M5_person:Party(person)
M5_act:What(drive or operate mobile food preparation unit)
M5_unit:SemanticEntity(mobile food preparation unit)
M5_locgroup:LogicalGroup(public street or private property; operator_type=OR)
M5_street:Location(public street)
M5_property:Location(private property)
M5_cond:ConditionGroup(all persons in vehicle are seated)
M5_exception:Defeasibility(seated-person exception; defeasibility_type=exception)
\end{olgppgraph}

\noindent\textbf{Edges.}
\begin{olgppgraph}
(M5_o)-[:DEONTIC_MODALITY {modality=prohibition}]->(M5_person)
(M5_o)-[:WHAT_REL]->(M5_act)
(M5_act)-[:PERFORMED_BY]->(M5_person)
(M5_act)-[:RELATED_TO]->(M5_unit)
(M5_act)-[:SPATIAL]->(M5_locgroup)
(M5_locgroup)-[:OR]->(M5_street)
(M5_locgroup)-[:OR]->(M5_property)
(M5_exception)-[:EXCEPTION]->(M5_o)
(M5_exception)-[:CONTEXTUAL]->(M5_cond)
\end{olgppgraph}

\noindent\textbf{Annotation note.} The ``unless'' clause is modeled as an exception to the driving/operating prohibition.

\subsubsection*{M6. Section 6.02.040(B): Cooking/Food Preparation While Operating}
\noindent\textbf{Provision unit.} No person may drive or operate a mobile food preparation unit on public street or private property while cooking or food preparation is going on in the vehicle.

\noindent\textbf{Nodes.}
\begin{olgppgraph}
M6_o:ObligationTrigger(operate while cooking prohibition; modality=prohibition)
M6_person:Party(person)
M6_operate:What(drive or operate mobile food preparation unit)
M6_cook:What(cooking or food preparation in vehicle)
M6_vehicle:SemanticEntity(vehicle/mobile food preparation unit)
M6_locgroup:LogicalGroup(public street or private property; operator_type=OR)
M6_during:Timex(while cooking/food preparation is going on; temporal_type=overlap)
\end{olgppgraph}

\noindent\textbf{Edges.}
\begin{olgppgraph}
(M6_o)-[:DEONTIC_MODALITY {modality=prohibition}]->(M6_person)
(M6_o)-[:WHAT_REL]->(M6_operate)
(M6_operate)-[:PERFORMED_BY]->(M6_person)
(M6_operate)-[:SPATIAL]->(M6_locgroup)
(M6_cook)-[:RELATED_TO]->(M6_vehicle)
(M6_operate)-[:DURING]->(M6_during)
(M6_operate)-[:TEMPORAL_OVERLAP]->(M6_cook)
(M6_locgroup)-[:OR]->(public_street)
(M6_locgroup)-[:OR]->(private_property)
\end{olgppgraph}

\noindent\textbf{Annotation note.} This provision is valuable because the temporal condition is an activity overlap, not a clock time.

\subsection{Single-Use Plastic Foodware Regulations: Chapter 6.20}

\subsubsection*{F1. Section 6.20.020: Food Service Provider Definition}
\noindent\textbf{Provision unit.} A food service provider includes restaurants, similar food facilities, mobile food facilities, mobile food vendors, food trucks, temporary food facilities, transient lodging facilities that provide prepared food, and specified statutory entities; exempt entities under Section 6.20.070 are excluded.

\noindent\textbf{Nodes.}
\begin{olgppgraph}
F1_fsp:SemanticEntity(food service provider; semantic_type=defined_term)
F1_members:LogicalGroup(enumerated provider classes; operator_type=OR)
F1_restaurant:SemanticEntity(restaurant/cafe/coffee shop/fast-food/etc.)
F1_mobile:SemanticEntity(mobile food facility/mobile vendor/food truck/temporary food facility)
F1_lodging:SemanticEntity(transient lodging facility providing prepared food)
F1_statutory:Reference(California Health and Safety Code 113789(a),(b))
F1_exempt:Reference(Section 6.20.070 excluded entities)
F1_d:Defeasibility(excluded entities; defeasibility_type=exception)
\end{olgppgraph}

\noindent\textbf{Edges.}
\begin{olgppgraph}
(F1_restaurant)-[:SUBCLASS_OF]->(F1_fsp)
(F1_mobile)-[:SUBCLASS_OF]->(F1_fsp)
(F1_lodging)-[:SUBCLASS_OF]->(F1_fsp)
(F1_statutory)-[:SUBJECT_TO]->(F1_fsp)
(F1_members)-[:OR]->(F1_restaurant/F1_mobile/F1_lodging/F1_statutory)
(F1_d)-[:EXCEPTION]->(F1_fsp)
(F1_d)-[:SUBJECT_TO]->(F1_exempt)
(F1_exempt)-[:MUTUAL_EXCLUSIVITY]->(F1_fsp)
\end{olgppgraph}

\noindent\textbf{Annotation note.} This definition uses taxonomy, OR enumeration, external statutory reference, and exclusion.

\subsubsection*{F2. Section 6.20.020: Compostable Definition}
\noindent\textbf{Provision unit.} Compostable materials must satisfy all listed conditions: city organics-program acceptance, ASTM compostability standard, labeling requirements, and PFAS standards.

\noindent\textbf{Nodes.}
\begin{olgppgraph}
F2_comp:SemanticEntity(compostable; semantic_type=defined_term)
F2_cond:ConditionGroup(compostability conditions; group_type=definition_condition)
F2_and:LogicalGroup(all compostability conditions; operator_type=AND)
F2_citymgr:Party(City Manager or designee)
F2_organics:Reference(city organic materials collection program)
F2_astm:Reference(ASTM standard specification / PRC 42356)
F2_label:Reference(PRC 42357 labeling requirements)
F2_pfas:Reference(HSC 109000 PFAS standards)
\end{olgppgraph}

\noindent\textbf{Edges.}
\begin{olgppgraph}
(F2_comp)-[:CONTEXTUAL]->(F2_cond)
(F2_cond)-[:MEMBER]->(F2_and)
(F2_and)-[:AND]->(F2_organics)
(F2_and)-[:AND]->(F2_astm)
(F2_and)-[:AND]->(F2_label)
(F2_and)-[:AND]->(F2_pfas)
(F2_citymgr)-[:AGENCY {scope=determine_accepted_collection}]->(F2_organics)
(F2_comp)-[:SUBJECT_TO]->(F2_astm/F2_label/F2_pfas)
\end{olgppgraph}

\noindent\textbf{Annotation note.} This is a clean example of a conjunctive constitutive definition.

\subsubsection*{F3. Section 6.20.020: Foodware and Foodware Accessory Definitions}
\noindent\textbf{Provision unit.} Foodware includes items used for containing, serving, or consuming prepared food, including containers and foodware accessories; foodware excludes egg cartons, meat trays, coolers, ice chests, and packing materials. Foodware accessories include utensils, straws, stirrers, condiments, cup lids/sleeves, and similar items used as part of or alongside prepared food.

\noindent\textbf{Nodes.}
\begin{olgppgraph}
F3_foodware:SemanticEntity(foodware; semantic_type=defined_term)
F3_accessory:SemanticEntity(foodware accessory; semantic_type=defined_term)
F3_prepared:SemanticEntity(prepared food)
F3_included:LogicalGroup(containers/cups/bowls/plates/trays/cartons/boxes/accessories; operator_type=OR)
F3_excluded:LogicalGroup(egg cartons/meat trays/coolers/ice chests/packing materials; operator_type=OR)
F3_formula:Formula(foodware minus excluded items; operation=set_difference)
\end{olgppgraph}

\noindent\textbf{Edges.}
\begin{olgppgraph}
(F3_accessory)-[:SUBCLASS_OF]->(F3_foodware)
(F3_included)-[:OR]->(container/cup/bowl/plate/tray/carton/box/F3_accessory)
(F3_included)-[:PART_OF {role=included_items}]->(F3_foodware)
(F3_excluded)-[:OR]->(egg_carton/meat_tray/cooler/ice_chest/packing_material)
(F3_excluded)-[:MUTUAL_EXCLUSIVITY]->(F3_foodware)
(F3_formula)-[:SUBTRACTION]->(F3_excluded)
(F3_accessory)-[:USED_ALONGSIDE {declared=true}]->(F3_prepared)
(F3_accessory)-[:RELATED_TO]->(F3_prepared)
\end{olgppgraph}

\noindent\textbf{Annotation note.} Definition exclusions are represented using both mutual exclusivity and set-difference formula structure.

\subsubsection*{F4. Section 6.20.030(A--B): Foodware Prohibition and Reusable/Compostable Requirement}
\noindent\textbf{Provision unit.} Food service providers may not provide prepared food in polystyrene foodware and must use only reusable or compostable foodware for on-premises and off-premises consumption and all ordering methods.

\noindent\textbf{Nodes.}
\begin{olgppgraph}
F4_proh:ObligationTrigger(polystyrene foodware prohibition; modality=prohibition)
F4_obl:ObligationTrigger(reusable or compostable requirement; modality=obligation)
F4_fsp:Party(food service provider)
F4_provide:What(provide prepared food in foodware)
F4_polystyrene:SemanticEntity(polystyrene foodware)
F4_reusable:SemanticEntity(reusable foodware)
F4_compostable:SemanticEntity(compostable foodware)
F4_method:LogicalGroup(on/off premises and ordering methods; operator_type=OR)
\end{olgppgraph}

\noindent\textbf{Edges.}
\begin{olgppgraph}
(F4_proh)-[:DEONTIC_MODALITY {modality=prohibition}]->(F4_fsp)
(F4_proh)-[:WHAT_REL]->(F4_provide)
(F4_provide)-[:RELATED_TO]->(F4_polystyrene)
(F4_obl)-[:DEONTIC_MODALITY {modality=obligation}]->(F4_fsp)
(F4_obl)-[:WHAT_REL]->(F4_provide)
(F4_provide)-[:CONTEXTUAL]->(F4_method)
(allowed_foodware)-[:OR]->(F4_reusable)
(allowed_foodware)-[:OR]->(F4_compostable)
(F4_polystyrene)-[:NOT]->(allowed_foodware)
\end{olgppgraph}

\noindent\textbf{Annotation note.} This provision pairs a prohibition with a positive obligation.

\subsubsection*{F5. Section 6.20.030(C): Foil Wrapper and Straw Exceptions}
\noindent\textbf{Provision unit.} Notwithstanding the foodware prohibition, non-compostable foil wrappers may be used if necessary and recyclable, and a small supply of single-use plastic straws may be maintained for consumers with disabilities or medical/health conditions upon request.

\noindent\textbf{Nodes.}
\begin{olgppgraph}
F5_foil:Defeasibility(foil-wrapper exception; defeasibility_type=exception)
F5_straw:Defeasibility(straw exception; defeasibility_type=exception)
F5_foilitem:SemanticEntity(non-compostable foil wrapper)
F5_strawitem:SemanticEntity(single-use plastic straw)
F5_condfoil:ConditionGroup(necessary to contain/form food and accepted in recycling)
F5_condstraw:ConditionGroup(disability/medical/health condition and consumer request)
F5_req:What(request single-use plastic straw)
F5_consumer:Party(consumer with disability/medical/health condition)
\end{olgppgraph}

\noindent\textbf{Edges.}
\begin{olgppgraph}
(F5_foil)-[:EXCEPTION]->(F4_proh)
(F5_foil)-[:CONTEXTUAL]->(F5_condfoil)
(F5_condfoil)-[:AND]->(necessary_to_contain_or_form_food)
(F5_condfoil)-[:AND]->(accepted_in_recycling_program)
(F5_straw)-[:EXCEPTION]->(F4_proh)
(F5_straw)-[:CONTEXTUAL]->(F5_condstraw)
(F5_req)-[:PERFORMED_BY]->(F5_consumer)
(provide_straw)-[:PREREQUISITE]->(F5_req)
(F5_strawitem)-[:SUBJECT_TO]->(Section_6_20_070)
(F5_strawitem)-[:SUBJECT_TO]->(Section_6_20_050)
\end{olgppgraph}

\noindent\textbf{Annotation note.} ``Notwithstanding'' is modeled as an explicit exception relation.

\subsubsection*{F6. Section 6.20.040: City Facilities and City-Affiliated Events}
\noindent\textbf{Provision unit.} Procurement, use, or distribution of noncompliant foodware is prohibited at city facilities and city-affiliated events; city departments, agents, employees, and designees may not procure noncompliant foodware; permits, leases, rentals, and contracts must require applicants to assume compliance responsibility.

\noindent\textbf{Nodes.}
\begin{olgppgraph}
F6_proh:ObligationTrigger(city facility/event foodware prohibition; modality=prohibition)
F6_citygroup:PartyGroup(city departments/agents/employees/designees; is_collective=true)
F6_city:Party(City of Carlsbad)
F6_facility:SpecificLocation(city facilities)
F6_event:EventTrigger(city-affiliated event)
F6_contract:Reference(special event permit/rental/lease/vendor contract/approval)
F6_responsibility:ObligationTrigger(applicant assumes compliance responsibility; modality=obligation)
F6_applicant:Party(applicant)
\end{olgppgraph}

\noindent\textbf{Edges.}
\begin{olgppgraph}
(F6_citygroup)-[:HAS_MEMBER]->(departments/agents/employees/designees)
(departments/agents/employees/designees)-[:AGENCY]->(F6_city)
(F6_proh)-[:DEONTIC_MODALITY {modality=prohibition}]->(F6_citygroup)
(F6_proh)-[:SPATIAL]->(F6_facility)
(F6_proh)-[:CONTEXTUAL]->(F6_event)
(F6_facility)-[:HAS_JURISDICTION]->(City_of_Carlsbad)
(F6_contract)-[:PREREQUISITE]->(F6_responsibility)
(F6_responsibility)-[:DEONTIC_MODALITY {modality=obligation}]->(F6_applicant)
(F6_applicant)-[:DELEGATION {scope=compliance_responsibility}]->(F6_contract)
\end{olgppgraph}

\noindent\textbf{Annotation note.} This provision uses party groups, agency, delegation/responsibility, and public-facility/event scope.

\subsubsection*{F7. Section 6.20.050: Foodware Accessories Upon Request}
\noindent\textbf{Provision unit.} Regulated entities may not provide single-use foodware accessories or standard condiments packaged for single use unless requested by the consumer, subject to listed exceptions for drive-through, airport, delivery, self-service dispensers, and related cases.

\noindent\textbf{Nodes.}
\begin{olgppgraph}
F7_proh:ObligationTrigger(accessory provision prohibition absent request; modality=prohibition)
F7_entity:Party(regulated entity)
F7_consumer:Party(consumer)
F7_provide:What(provide single-use accessory or condiment)
F7_request:EventTrigger(consumer request)
F7_cond:ConditionGroup(request condition)
F7_exceptions:LogicalGroup(E through H exceptions; operator_type=OR)
F7_platform:Party(third-party food delivery platform)
\end{olgppgraph}

\noindent\textbf{Edges.}
\begin{olgppgraph}
(F7_proh)-[:DEONTIC_MODALITY {modality=prohibition}]->(F7_entity)
(F7_proh)-[:WHAT_REL]->(F7_provide)
(F7_provide)-[:PREREQUISITE]->(F7_request)
(F7_request)-[:PERFORMED_BY]->(F7_consumer)
(F7_request)-[:IF_TRUE]->(provide_accessory_permission)
(F7_request)-[:IF_FALSE]->(F7_proh)
(F7_exceptions)-[:OR]->(drive_through/airport/lids_sleeves/platform_options/self_service)
(F7_exceptions)-[:EXCEPTION]->(F7_proh)
(platform_option_obligation)-[:DEONTIC_MODALITY {modality=obligation}]->(F7_platform)
\end{olgppgraph}

\noindent\textbf{Annotation note.} This provision exercises request-as-prerequisite, exceptions, platform obligations, and if-false interpretation.

\subsubsection*{F8. Section 6.20.050(A): More Stringent Local Rule Governs}
\noindent\textbf{Provision unit.} Regulated entities must distribute accessories in accordance with Chapter 6.20 and AB 1276; to the extent Chapter 6.20 is more stringent, Chapter 6.20 governs as permitted by state law.

\noindent\textbf{Nodes.}
\begin{olgppgraph}
F8_local:Reference(Chapter 6.20)
F8_state:Reference(AB 1276 / PRC 42271(h))
F8_rule:ObligationTrigger(accessory distribution compliance; modality=obligation)
F8_entity:Party(regulated entity)
F8_cond:ConditionGroup(local chapter more stringent)
F8_def:Defeasibility(local rule priority; defeasibility_type=precedence)
\end{olgppgraph}

\noindent\textbf{Edges.}
\begin{olgppgraph}
(F8_rule)-[:DEONTIC_MODALITY {modality=obligation}]->(F8_entity)
(F8_rule)-[:SUBJECT_TO]->(F8_local)
(F8_rule)-[:SUBJECT_TO]->(F8_state)
(F8_def)-[:PRECEDENCE {basis=more_stringent}]->(F8_local)
(F8_local)-[:OVERRIDE {condition=more_stringent}]->(F8_state)
(F8_def)-[:CONTEXTUAL]->(F8_cond)
\end{olgppgraph}

\noindent\textbf{Annotation note.} This is a precedence/override relation between local and state-law requirements, not merely an exception.

\subsubsection*{F9. Section 6.20.060(E): Educational Materials Posting}
\noindent\textbf{Provision unit.} Within 30 days of the effective date, food service providers must post city-provided educational materials near the menu, point-of-sale counter, or other clearly visible location prior to ordering.

\noindent\textbf{Nodes.}
\begin{olgppgraph}
F9_o:ObligationTrigger(post educational materials; modality=obligation)
F9_fsp:Party(food service provider)
F9_materials:SemanticEntity(educational materials)
F9_cityevent:EventTrigger(city provides materials)
F9_deadline:Timex(30 days after effective date)
F9_visible:LocationPredicate(menu/POS/visible location prior to ordering)
F9_order:EventTrigger(consumer ordering)
\end{olgppgraph}

\noindent\textbf{Edges.}
\begin{olgppgraph}
(F9_o)-[:DEONTIC_MODALITY {modality=obligation}]->(F9_fsp)
(F9_o)-[:WHAT_REL]->(post_materials)
(post_materials)-[:RELATED_TO]->(F9_materials)
(F9_cityevent)-[:IF_TRUE]->(F9_o)
(F9_o)-[:TEMPORAL_SEQUENCE {max_gap=30_days}]->(F9_deadline)
(F9_deadline)-[:AFTER]->(effective_date)
(post_materials)-[:LOCATION_PREDICATE]->(F9_visible)
(F9_visible)-[:BEFORE]->(F9_order)
\end{olgppgraph}

\noindent\textbf{Annotation note.} This combines an if-true trigger, deadline, spatial visibility, and prior-to-ordering temporal relation.

\subsubsection*{F10. Section 6.20.070: Exemptions and Waivers}
\noindent\textbf{Provision unit.} Certain institutions are excluded; the City Manager may temporarily exempt entities during an emergency; may exempt items when no feasible alternative exists; disability straw requests are preserved; temporary waivers may last up to six months.

\noindent\textbf{Nodes.}
\begin{olgppgraph}
F10_excluded:Defeasibility(excluded institutions; defeasibility_type=exception)
F10_emergency:EventTrigger(emergency)
F10_emergencyex:Defeasibility(emergency temporary exemption)
F10_feasible:ConditionGroup(no reasonably feasible alternative)
F10_itemex:Defeasibility(item feasibility exemption)
F10_manager:Party(City Manager or designee)
F10_waiver:Defeasibility(full or partial temporary waiver)
F10_six:Timex(up to six months)
F10_straw:Defeasibility(disability/medical straw request preservation)
\end{olgppgraph}

\noindent\textbf{Edges.}
\begin{olgppgraph}
(F10_excluded)-[:EXCEPTION]->(Chapter_6_20_requirements)
(F10_emergency)-[:IF_TRUE]->(F10_emergencyex)
(F10_emergencyex)-[:EXCEPTION]->(Chapter_6_20_requirements)
(F10_manager)-[:AGENCY {scope=grant_exemption}]->(F10_emergencyex)
(F10_itemex)-[:CONTEXTUAL]->(F10_feasible)
(F10_itemex)-[:EXCEPTION]->(foodware_requirements)
(F10_waiver)-[:DURING]->(F10_six)
(F10_waiver)-[:TEMPORAL_MODIFIER]->(F10_six)
(F10_manager)-[:AGENCY {scope=adopt_waiver_rules}]->(F10_waiver)
(F10_straw)-[:EXCEPTION]->(single_use_plastic_straw_restriction)
\end{olgppgraph}

\noindent\textbf{Annotation note.} This provision exercises multiple exception mechanisms and agency authority.

\subsubsection*{F11. Section 6.20.080: Enforcement, Records, Inspection, and Penalty Cap}
\noindent\textbf{Provision unit.} Enforcement begins on specified dates; regulated entities must maintain compliance records and make them available for inspection; subsequent violations are infractions with a $25 per day fine capped at $300 annually; remedies are cumulative.

\noindent\textbf{Nodes.}
\begin{olgppgraph}
F11_agency:Party(enforcement agency/official)
F11_dates:Timex(June 1, 2022; July 1, 2023)
F11_records:ObligationTrigger(maintain and make records available; modality=obligation)
F11_entity:Party(regulated entity)
F11_inspect:What(inspect records/premises)
F11_violation:EventTrigger(subsequent violation)
F11_penalty:Formula($25/day not to exceed $300/year; operation=min/multiply/max)
F11_daily:Amount($25 per day)
F11_cap:Amount($300 annually)
\end{olgppgraph}

\noindent\textbf{Edges.}
\begin{olgppgraph}
(enforce_accessory_rules)-[:ON]->(June_1_2022)
(enforce_other_rules)-[:ON]->(July_1_2023)
(F11_agency)-[:AGENCY]->(enforcement_powers)
(F11_records)-[:DEONTIC_MODALITY {modality=obligation}]->(F11_entity)
(F11_inspect)-[:PERFORMED_BY]->(F11_agency)
(F11_violation)-[:IF_TRUE]->(infraction)
(F11_penalty)-[:MULTIPLICATION]->(F11_daily)
(F11_penalty)-[:MAXIMUM]->(F11_cap)
(F11_penalty)-[:FORMULA_REL]->(days_in_violation)
(remedies)-[:RELATED_TO {status=cumulative}]->(penalties)
\end{olgppgraph}

\noindent\textbf{Annotation note.} This provision covers enforcement start dates, records/inspection obligations, and capped recurring penalties.

\subsection{Smoking in Unenclosed Dining Areas: Chapter 6.14}

\subsubsection*{S1. Sections 6.14.020--6.14.040: Smoking Prohibition and Reasonable Distance}
\noindent\textbf{Provision unit.} Smoking is prohibited in unenclosed dining areas within the City of Carlsbad and within a reasonable distance, defined as 20 feet in any direction, from such areas.

\noindent\textbf{Nodes.}
\begin{olgppgraph}
S1_o:ObligationTrigger(smoking prohibition; modality=prohibition)
S1_smoke:What(smoking; action_type=smoking)
S1_person:Party(any person)
S1_dining:Location(unenclosed dining area)
S1_city:Jurisdiction(City of Carlsbad)
S1_distance:Amount(20 feet; value=20, unit=feet)
S1_predicate:LocationPredicate(within reasonable distance in any direction)
\end{olgppgraph}

\noindent\textbf{Edges.}
\begin{olgppgraph}
(S1_o)-[:DEONTIC_MODALITY {modality=prohibition}]->(S1_person)
(S1_o)-[:WHAT_REL]->(S1_smoke)
(S1_smoke)-[:SPATIAL]->(S1_dining)
(S1_o)-[:HAS_JURISDICTION]->(S1_city)
(S1_smoke)-[:PROXIMITY_TO]->(S1_dining)
(S1_smoke)-[:AMOUNT]->(S1_distance)
(S1_smoke)-[:LOCATION_PREDICATE]->(S1_predicate)
\end{olgppgraph}

\noindent\textbf{Annotation note.} This provision is the cleanest example of a proximity constraint.

\subsubsection*{S2. Section 6.14.030: State/Federal Law Applies Where Already Prohibited}
\noindent\textbf{Provision unit.} Smoking is prohibited in unenclosed dining areas except places where smoking is already prohibited by state or federal law, in which case those laws apply.

\noindent\textbf{Nodes.}
\begin{olgppgraph}
S2_local:Reference(Chapter 6.14 local smoking prohibition)
S2_statefed:Reference(state or federal smoking law)
S2_cond:ConditionGroup(place already prohibited by state/federal law)
S2_def:Defeasibility(state/federal law governs; defeasibility_type=precedence)
\end{olgppgraph}

\noindent\textbf{Edges.}
\begin{olgppgraph}
(S2_local)-[:SUBJECT_TO]->(S2_statefed)
(S2_statefed)-[:PRECEDENCE]->(S2_local)
(S2_statefed)-[:OVERRIDE {condition=already_prohibited}]->(S2_local)
(S2_def)-[:CONTEXTUAL]->(S2_cond)
(S2_cond)-[:HAS_JURISDICTION]->(state_or_federal_scope)
\end{olgppgraph}

\noindent\textbf{Annotation note.} This is modeled as precedence/override, not an ordinary exception, because another legal source governs.

\subsubsection*{S3. Section 6.14.050: Optional Prohibition by Property Controller}
\noindent\textbf{Provision unit.} A person, corporation, legal entity, or employer with legal control over property may prohibit smoking on any part of that property even if smoking is not otherwise prohibited by law.

\noindent\textbf{Nodes.}
\begin{olgppgraph}
S3_perm:ObligationTrigger(optional smoking prohibition authority; modality=permission)
S3_controller:Party(property controller; party_type=person/corporation/legal_entity/employer)
S3_property:Location(controlled property)
S3_prohibit:What(prohibit smoking)
S3_cond:ConditionGroup(legal control over property)
\end{olgppgraph}

\noindent\textbf{Edges.}
\begin{olgppgraph}
(S3_perm)-[:DEONTIC_MODALITY {modality=permission}]->(S3_controller)
(S3_perm)-[:WHAT_REL]->(S3_prohibit)
(S3_prohibit)-[:SPATIAL]->(S3_property)
(S3_controller)-[:AGENCY {scope=legal_control}]->(S3_property)
(S3_perm)-[:CONTEXTUAL]->(S3_cond)
(S3_controller)-[:HAS_JURISDICTION {scope=private_property_control}]->(S3_property)
\end{olgppgraph}

\noindent\textbf{Annotation note.} This is a permission/entitlement-like rule giving private controllers authority to impose stricter local restrictions.

\subsubsection*{S4. Sections 6.14.060--6.14.070: No-Smoking Signs and Ashtrays}
\noindent\textbf{Provision unit.} Controllers of unenclosed dining areas where smoking is prohibited must post clear signs at each ingress point and one conspicuous point; letters must be at least one inch high; presence or absence of signs is not a defense; ashtrays and smoking disposal receptacles may not be placed where smoking is prohibited.

\noindent\textbf{Nodes.}
\begin{olgppgraph}
S4_signobl:ObligationTrigger(post no-smoking signs; modality=obligation)
S4_controller:Party(property/controller of dining area)
S4_area:Location(unenclosed dining area)
S4_ingress:LocationPredicate(each point of ingress and conspicuous point)
S4_letter:Amount(one inch; value=1, unit=inch)
S4_nodef:Defeasibility(no-sign-defense preclusion; defeasibility_type=precedence)
S4_ashtray:SemanticEntity(ashtray/smoking disposal receptacle)
S4_ashproh:ObligationTrigger(no ashtrays in prohibited areas; modality=prohibition)
\end{olgppgraph}

\noindent\textbf{Edges.}
\begin{olgppgraph}
(S4_signobl)-[:DEONTIC_MODALITY {modality=obligation}]->(S4_controller)
(S4_signobl)-[:SPATIAL]->(S4_area)
(post_sign)-[:LOCATION_PREDICATE]->(S4_ingress)
(S4_ingress)-[:PART_OF]->(S4_area)
(sign_letters)-[:AMOUNT]->(S4_letter)
(sign_letters)-[:COMPARISON {operator=greater_than_or_equal}]->(S4_letter)
(S4_nodef)-[:PRECEDENCE]->(sign_based_defense)
(S4_nodef)-[:NOT]->(defense_based_on_presence_or_absence_of_signs)
(S4_ashproh)-[:WHAT_REL]->(place_ashtray)
(place_ashtray)-[:RELATED_TO]->(S4_ashtray)
(place_ashtray)-[:SPATIAL]->(areas_where_smoking_prohibited)
\end{olgppgraph}

\noindent\textbf{Annotation note.} This provision exercises sign-placement location predicates, amount comparisons, negated defense, and prohibition of related objects in prohibited areas.

\subsection{Summary of Appendix Coverage}

The 35 provision units above exercise all normalized OLG++ node types and all major edge categories used in the main text. The annotations also identify three declared domain-specific relations: \texttt{USED\_FOR}, \texttt{ENTRANCE\_LEADS\_TO}, and \texttt{USED\_ALONGSIDE}. These are handled through the declared extension mechanism rather than by adding new primitive schema elements.

\end{document}